\documentclass{article}

\usepackage[preprint]{neurips_2026}

\PassOptionsToPackage{off}{epstopdf}

\usepackage[utf8]{inputenc}
\usepackage[T1]{fontenc}
\usepackage{url}
\usepackage{booktabs}
\usepackage{amsfonts}
\usepackage{nicefrac}
\usepackage{microtype}
\usepackage{xcolor}
\usepackage{amsmath}
\usepackage{amsthm}
\usepackage{amssymb}
\usepackage{graphicx}
\usepackage{wrapfig}

\graphicspath{{./}}
\DeclareGraphicsExtensions{.pdf,.png,.jpg,.jpeg}
\usepackage{multirow}
\usepackage{adjustbox}
\usepackage[colorlinks,urlcolor=purple,linkcolor=purple,citecolor=purple]{hyperref}
\usepackage[caption=false,font=footnotesize]{subfig}

\newtheorem{definition}{Definition}
\newtheorem{problem}{Problem}
\newtheorem{remark}{Remark}

\title{Differentiable Koopman Operator for Contrastive Learning on Dynamic Graphs}

\author{
  \textbf{Md Abrar Jahin}$^\diamondsuit$ \quad
  \textbf{Taufikur Rahman Fuad}$^\clubsuit$ \quad
  \textbf{Md Rizwan Parvez}$^\spadesuit$ \\
  $^\diamondsuit$University of Southern California \quad $^\clubsuit$Islamic University of Technology \\
  $^\spadesuit$Qatar Computing Research Institute (QCRI)
}

\begin{document}

\maketitle

\begin{abstract}
Real-world interaction networks are inherently dynamic: edges form and dissolve as node behavior shifts over time. Most snapshot-based contrastive methods encode temporal dependencies implicitly in encoder weights, without an explicit model of how node representations evolve, making them brittle under distribution shifts. We propose KAIROS (Koopman-Aligned Invariant Representations for Open Dynamic Systems), a self-supervised framework that embeds a differentiable Koopman operator within a dynamic graph contrastive learning loop to linearize temporal evolution in the learned embedding space. A dual-view encoder pairs raw node features with a graph-diffused structural view and is optimized with multi-granularity contrastive objectives across temporal windows. For anomaly detection, KAIROS uses the Koopman prediction residual together with temporal inconsistency and local neighborhood deviation to separate irregular behavior from predictable graph evolution. Evaluated on nine dynamic graph benchmarks, KAIROS achieves state-of-the-art anomaly detection results on all nine datasets, with gains of up to 23.15 ROC-AUC points over prior work, while remaining competitive for unsupervised node classification. These results show that explicit dynamics modeling provides a scalable and effective inductive bias for temporal graph representation learning.
\end{abstract}

\section{Introduction}
\label{sec:intro}

Real-world relational systems such as financial transaction networks, social platforms, and citation graphs are inherently dynamic. Dynamic Graph Representation Learning (DGRL) seeks node embeddings that capture structural dependencies and temporal evolution, enabling downstream tasks such as node classification, link prediction, and anomaly detection. Graph Neural Networks~(GNNs) are well established for static graphs, but extending them to dynamic settings remains challenging.

Many dynamic-graph encoders model temporal evolution through architectural mechanisms and lack a transition operator that predicts how node embeddings evolve between snapshots. Snapshot-based models aggregate per-snapshot neighborhoods, while continuous-time models encode interactions through functional time encodings, causal anonymous walks, neighborhood/community influence aggregation, or frequency-domain temporal features~\cite{sankar2020dysat,xu2020inductive,wang2021inductive,liu2021inductive,tianFreeDyGFrequencyEnhanced2024}. These designs do not provide a parameterized one-step operator over node embeddings. Without such a model, predictable evolution and anomalous deviation are difficult to separate under distribution shift, where both may appear as movement in embedding space~\cite{zhangDynamicGraphNeural2022,zhangSpectralInvariantLearning2023,zhaoCausalityInspiredSpatialTemporalExplanations2024}.

This limitation is amplified in self-supervised dynamic graph learning. Strong dynamic contrastive baselines such as CLDG and CLDG++ maximize agreement between temporal views~\cite{CLDG,CLDGPP}. This provides a useful smoothness bias but does not model how representations evolve. When evolution is non-smooth, view agreement can penalize legitimate change and fail to isolate deviations. Existing dynamic graph anomaly detectors address temporal anomalies through dynamic embeddings, transformer-based spatiotemporal modeling, or semi-supervised pseudo-labeling~\cite{NetWalk,TADDY,SAD}. However, they do not define anomaly scores as residuals of a learned transition operator.

Koopman operator theory offers a direct way to impose such a transition model: observables of nonlinear dynamical systems evolve linearly in an infinite-dimensional lifted space~\cite{koopman1931hamiltonian,mezic2005spectral}. Neural Koopman approximations have been used for forecasting, control, and sequence modeling through reconstruction, forecasting, or operator-regression objectives~\cite{takeishiLearningKoopmanInvariant2017,DBLP:conf/icml/AzencotELM20,wangKoopmanNeuralOperator2023,hanDeSKOStabilityAssuredRobust2022,bermanMultifactorSequentialDisentanglement2023}. To our knowledge, no prior work couples a graph encoder and Koopman operator end-to-end under a contrastive objective, or uses the resulting Koopman residual as an unsupervised anomaly signal on dynamic graphs.

We propose \textbf{KAIROS} (\textbf{K}oopman-\textbf{A}ligned \textbf{I}nvariant \textbf{R}epresentations for \textbf{O}pen Dynamic \textbf{S}ystems), a self-supervised framework that embeds a differentiable Koopman operator inside a dynamic graph contrastive learning loop. A dual-view GNN encoder processes raw features and a Personalized PageRank (PPR)-diffused view~\cite{appnp}, and optimizes an InfoNCE objective across temporal windows. A linear operator $\mathbf{K}$, initialized at the identity and regularized toward orthogonality, is co-optimized to predict each node's next-snapshot embedding. Initializing $\mathbf{K}=\mathbf{I}$ makes the identity transition a special case of the learned operator, so any deviation from identity is loss-driven. The same encoder supports two downstream tasks without architectural change: (i) for unsupervised node classification, the Koopman regularizer is disabled and the frozen encoder feeds a linear probe; (ii) for unsupervised dynamic-graph anomaly detection, the Koopman prediction residual is combined with temporal inconsistency and local neighborhood deviation, each $z$-normalized, into a unified anomaly score.

Our main contributions are:
\textbf{(i)} We couple a differentiable Koopman operator with self-supervised dynamic graph contrastive learning, jointly optimizing the graph encoder and linear temporal operator end-to-end.
\textbf{(ii)} We introduce a dual-view contrastive objective over raw and PPR-diffused features, augmented with Koopman prediction and orthogonality regularization.
\textbf{(iii)} We define a fully unsupervised anomaly score from temporal inconsistency, Koopman prediction residual, and neighborhood deviation, without pseudo-labels or reconstruction losses.

\section{Related Work}
\label{sec:related}

\textbf{Dynamic Graph Representation Learning. }
Two regimes dominate DGRL. Discrete-time snapshot models such as DySAT~\cite{sankar2020dysat} aggregate per-snapshot neighborhoods and combine them with temporal self-attention, while continuous-time models such as TGAT~\cite{xu2020inductive}, CAW~\cite{wang2021inductive}, MNCI~\cite{liu2021inductive}, and FreeDyG~\cite{tianFreeDyGFrequencyEnhanced2024} encode interactions through functional time encodings, causal anonymous walks, neighborhood/community influence aggregation, or frequency-domain temporal features. A complementary line improves the expressivity and stability of temporal message passing through long-range propagation and refined inductive biases~\cite{gravinaLongRangePropagation2024,yuBetterDynamicGraph2023,heegUsingTimeAwareGraph2024}. None of these designs supplies an \emph{explicit model of how representations evolve} between snapshots: temporal dependencies are absorbed into encoder weights, leaving distribution shift as a recurring failure mode~\cite{zhangDynamicGraphNeural2022,zhangSpectralInvariantLearning2023,zhaoCausalityInspiredSpatialTemporalExplanations2024}. The limitation is representational rather than architectural: predictable evolution and anomalous deviation become indistinguishable. KAIROS departs from this setting by coupling representation learning with a differentiable Koopman operator that explicitly parameterizes inter-snapshot dynamics, turning temporal predictability into an inductive bias.

\textbf{Self-Supervised and Contrastive Learning on Graphs. }
Static graph contrastive learning is mature, with a well-mapped design space~\cite{velickovic2019deep,hassani2020contrastive,zhu2020deep,youGraphContrastiveLearning2020,zhang2021canonical,zhengRethinkingScalingGraph2022} that modifies topology or features and aligns the resulting views under InfoNCE or redundancy-reduction objectives~\cite{zbontarBarlowTwinsSelfSupervised2021,oord2018representation}. Subsequent analyses have exposed structural fragilities of this framework: asymmetric encoders without augmentations match or exceed augmentation-heavy baselines~\cite{xiaoSimpleAsymmetricGraph2023}; perfect alignment can be harmful under topology shifts~\cite{liuPerfectAlignmentMay2024}; hidden architectural mechanisms often dominate loss-level behavior~\cite{guoArchitectureMattersUncovering2023,heExploitationLatentMechanism2024}; and InfoNCE admits a spectral-clustering interpretation~\cite{tanContrastiveLearningSpectral2024,koromilasBridgingMiniBatchAsymptotic2024}. A complementary line studies spectral, principled, or efficiency-aware augmentations~\cite{linSpectralAugmentationSelfSupervised2023,yuProvableTrainingGraph2023,xiaoEfficientContrastiveLearning2024,zhuoUnifiedGraphAugmentations2024,wangUncoveringRedundancyGraph2024}. When extended to dynamic graphs, however, these advances are largely stripped of their temporal premise: CLDG~\cite{CLDG} and CLDG++~\cite{CLDGPP} treat temporally sampled subgraphs as augmented views and maximize InfoNCE agreement across time windows, optionally at multiple time spans, but impose no explicit dynamics between those windows. The implicit assumption that temporally adjacent embeddings remain \emph{similar} reduces to a smoothness penalty that becomes brittle when graph evolution is non-smooth.

\textbf{Graph Anomaly Detection. }
Unsupervised anomaly detection on attributed graphs is dominated by context--patch disagreement methods that combine contrastive or reconstruction objectives with multi-scale or augmentation-aware designs~\cite{CoLA,ANEMONE,GRADATE,AGEIS,tangRethinkingGraphNeural2022,linUniGADUnifyingMultilevel2024}. These methods are powerful but inherently static: they ignore temporal evolution and repeatedly flag the same structural anomaly at every snapshot, reducing precision under non-stationary baselines. Contrastive graph anomaly detection refines this framework along three axes: structural imbalance~\cite{xuRevisitingGraphContrastive2025}, clean-view formulations~\cite{jinRethinkingContrastiveLearning2025}, and dual-encoder augmentation~\cite{wuDualEncoderContrastive2025}. Existing dynamic-graph detectors fail in different ways: NetWalk~\cite{NetWalk} relies on shallow random-walk embeddings and lacks attribute-aware expressiveness; TADDY~\cite{TADDY} applies self-attention over edge-centric temporal subgraphs whose complexity grows quadratically, limiting scalability on realistic graphs; SAD~\cite{SAD} is semi-supervised and depends on pseudo-labels derived from partial ground truth, making it unsuitable for the fully unsupervised setting; and spatiotemporal anomaly networks~\cite{xuDynamicMultiscaleSpatiotemporal2026,yuTemporalMultivariatefactorsIndependence2025} target multivariate time series rather than evolving topologies.

\textbf{Koopman Operator Theory in Machine Learning. }
Koopman operator theory provides a linear representation of nonlinear dynamics in an infinite-dimensional space~\cite{koopman1931hamiltonian,mezic2005spectral}, and its finite-dimensional approximations now support a broad class of learning methods. Foundational work established neural realizations of Koopman embeddings~\cite{takeishiLearningKoopmanInvariant2017,DBLP:conf/icml/AzencotELM20,tayalKoopmanInvertibleAutoencoder2023,frionAugmentedInvertibleKoopman2025,jinExtendedDynamicMode2024} and rigorous statistical guarantees~\cite{kosticLearningDynamicalSystems2022,kosticSharpSpectralRates2023,houSparseLearningDynamical2023}. Application-driven extensions span operator forecasting~\cite{wangKoopmanNeuralOperator2023}, robust control~\cite{hanDeSKOStabilityAssuredRobust2022}, action recognition~\cite{wangNeuralKoopmanPooling2023}, sequence disentanglement~\cite{bermanMultifactorSequentialDisentanglement2023}, structured operator parameterizations~\cite{zhangSKOLRStructuredKoopman2025}, and recent neural-ODE and probabilistic forecasting hybrids~\cite{baiKoNODEKoopmanDrivenNeural2025,akaBalancedNeuralODEs2025,zhengKooNProVarianceAwareKoopman2025,wuK2VAEKoopmanKalmanEnhanced2025}. On graphs, however, Koopman theory remains underexplored. The closest related work is Shi et al.~\cite{shiWhenGraphNeural2025}, which links GNN message passing to dynamic mode decomposition for supervised dynamics modeling, without a contrastive or anomaly-detection component. KAIROS instead uses a differentiable Koopman operator inside self-supervised dynamic graph contrastive learning and uses its prediction residual as an unsupervised anomaly signal. 

\section{Background and Preliminaries}
\label{sec:prelim}

We consider dynamic graphs represented as temporal snapshots
$\{\mathcal{G}^1,\ldots,\mathcal{G}^T\}$ over a fixed node universe
$\mathcal{V}$, with node features $\mathbf{X}$ and timestamp function
$\mathcal{T}$. Continuous-time graphs are converted to snapshots by
partitioning the temporal span into $T$ equal-width windows. KAIROS uses
two views of each snapshot: the raw feature view $\mathbf{X}$ and a
PPR-diffused structural view $\mathbf{X}_{\mathrm{ppr}}$.

The goal is to learn a self-supervised encoder $f_\theta$ that supports
two downstream tasks: unsupervised dynamic node classification via a frozen
linear probe, and unsupervised dynamic graph anomaly detection via a scalar
node-level anomaly score. KAIROS models temporal evolution in the learned
embedding space with a linear Koopman operator $\mathbf{K}$, whose prediction
residual provides one anomaly signal. Details of this section is in Appendix!\ref{app:prelim}

\section{KAIROS}
\label{sec:kairos}

\begin{figure}[t]
\centering  
\includegraphics[width=\linewidth]{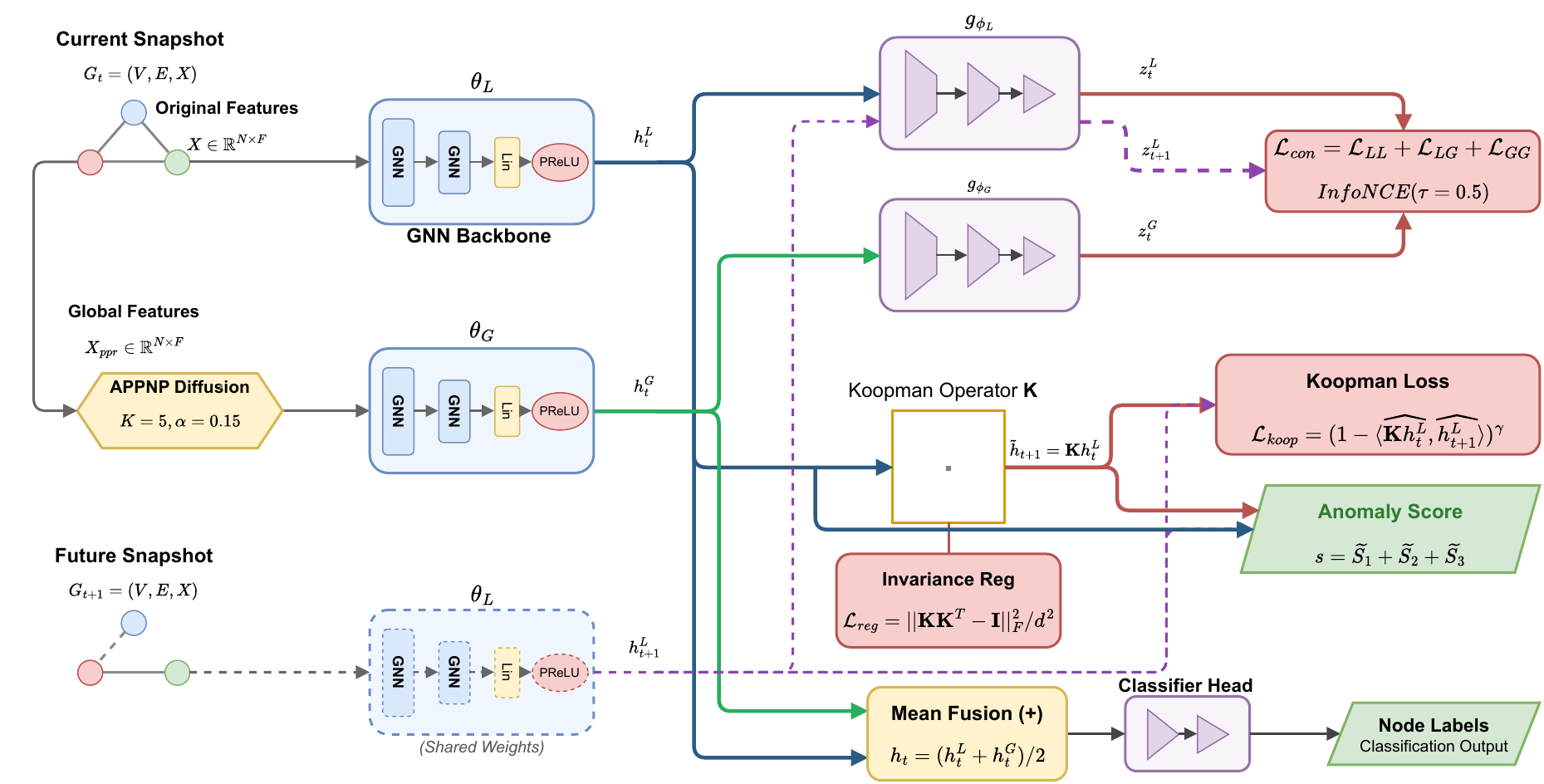}  
\caption{Overview of the KAIROS architecture.}
\label{fig:architecture}
\end{figure}

KAIROS is a self-supervised framework for learning node representations from dynamic graphs and addressing both Problems~\ref{prob:clf} and~\ref{prob:ano} with a shared encoder.
The framework consists of four coupled components:
\emph{(i)}~temporal snapshot construction and inductive view sampling (Section~\ref{subsec:views});
\emph{(ii)}~a dual-encoder architecture with independent GNN backbones for raw and PPR-diffused views, equipped with SimCLR\,v2 projection heads (Section~\ref{subsec:encoder});
\emph{(iii)}~a multi-granularity contrastive objective coupling local, global, and cross-view signals across time (Section~\ref{subsec:contrast});
\emph{(iv)}~a Koopman temporal operator $\mathbf{K}$ that linearizes embedding-space dynamics, regularizes training, and produces a per-node anomaly residual at inference (Section~\ref{subsec:koopman}).
At inference, three complementary anomaly signals are $z$-normalized and summed into a unified score (Section~\ref{subsec:anomaly}). An overview is shown in Figure~\ref{fig:architecture}.

\subsection{Temporal Graph Construction and View Generation}
\label{subsec:views}

\textbf{Snapshot and dual-view construction. }
KAIROS partitions the temporal edge stream into $T$ equal-width canonical snapshots and canonicalizes each snapshot by removing duplicate edges, symmetrizing the graph, and adding self-loops.
Because the benchmarks are timestamped edge lists without raw node attributes, KAIROS pairs sinusoidal node features $\mathbf{X}$ with a snapshot-specific PPR-diffused structural view $\mathbf{X}^k_{\mathrm{ppr}}$.
Full construction equations and APPNP parameters are in Appendix~\ref{app:snapshot_views}.

\textbf{Temporal view sampling.}
At each training step, KAIROS samples $V\le T$ temporal views and maps each to the nearest canonical snapshot index $\pi^{(v)}$. We use random, sequential, low-overlap, and high-overlap sampling as temporal augmentations; details are in Appendix~\ref{app:kairos_details}.

\textbf{Inductive anchor sampling.}
KAIROS samples up to $B{=}256$ anchor nodes from the source nodes shared across the sampled views. Small graphs use full-neighbor expansion, while larger graphs use multi-layer neighbor sampling with fanout $(20,20)$, bounding memory by the sampled computation graph rather than $|\mathcal{V}|$.

\subsection{Dual-Encoder Architecture}
\label{subsec:encoder}

\textbf{Encoder branches. }
KAIROS maintains two GNN backbones with \emph{disjoint} parameter sets:
\begin{equation}
  \mathbf{h}^{L,(v)}_i = \sigma_{\mathrm{PReLU}}\!\bigl(\mathbf{W}_L\,\Phi^L_{\theta_L}(\mathcal{G}^{\pi^{(v)}},\mathbf{X})_i\bigr),\qquad
  \mathbf{h}^{G,(v)}_i = \sigma_{\mathrm{PReLU}}\!\bigl(\mathbf{W}_G\,\Phi^G_{\theta_G}(\mathcal{G}^{\pi^{(v)}},\mathbf{X}^{\pi^{(v)}}_{\mathrm{ppr}})_i\bigr)
  \label{eq:enc}
\end{equation}
where $\Phi^L_{\theta_L}$ and $\Phi^G_{\theta_G}$ are drawn from $\{\text{GCN},\text{GAT},\text{GraphSAGE},\text{SGC},\text{H2GCN}\}$ (the same family selected per dataset on the validation set, but with independent weights), $\mathbf{W}_L,\mathbf{W}_G\in\mathbb{R}^{d\times d}$ are learnable post-aggregation projections, and $\sigma_{\mathrm{PReLU}}$ is a per-branch PReLU activation.
All backbones use $L{=}2$ message-passing layers, hidden dimension $128$, and output dimension $d{=}64$.
The pre-projection embeddings $\mathbf{h}^{L,(v)}_i$ and $\mathbf{h}^{G,(v)}_i$ are used for downstream evaluation; $\mathbf{h}^{L,(v)}_i$ alone feeds the Koopman temporal head (Section~\ref{subsec:koopman}), which linearizes the original-view trajectory. Projection-head and temperature details are in Appendix~\ref{app:proj_temp}; projection heads are discarded at inference.
Maintaining disjoint weights is essential: a shared encoder applied to both $\mathbf{X}$ and $\mathbf{X}_{\mathrm{ppr}}$ produces trivially correlated outputs because $\mathbf{X}_{\mathrm{ppr}}$ is a smooth neighborhood average of $\mathbf{X}$, collapsing the cross-view contrastive signal.

\subsection{Multi-Granularity Contrastive Objective}
\label{subsec:contrast}

For an ordered pair of temporal views $(u,v)$ with $u<v$, KAIROS contrasts the projected embedding matrices $\mathbf{z}^{L,(u)},\mathbf{z}^{G,(u)},\mathbf{z}^{L,(v)},\mathbf{z}^{G,(v)}\in\mathbb{R}^{B\times d}$ along three axes.
Let $\mathrm{NCE}_\tau(\mathbf{Z}_1,\mathbf{Z}_2)$ denote the symmetric InfoNCE objective between two $\ell_2$-normalized batches $\mathbf{Z}_1,\mathbf{Z}_2\in\mathbb{R}^{B\times d}$,
\begin{equation}
  \mathrm{NCE}_\tau(\mathbf{Z}_1,\mathbf{Z}_2)
  = -\frac{1}{B}\sum_{i=1}^{B}\Biggl[
  \log\frac{\exp(\langle\mathbf{z}_{1,i},\mathbf{z}_{2,i}\rangle/\tau)}
           {\sum_{j=1}^{B}\exp(\langle\mathbf{z}_{1,i},\mathbf{z}_{2,j}\rangle/\tau)}
  \,+\,
  \log\frac{\exp(\langle\mathbf{z}_{2,i},\mathbf{z}_{1,i}\rangle/\tau)}
            {\sum_{j=1}^{B}\exp(\langle\mathbf{z}_{2,i},\mathbf{z}_{1,j}\rangle/\tau)}
  \Biggr]
  \label{eq:nce}
\end{equation}
implemented with per-row logit shifting $\tilde{\ell}_{ij}=\ell_{ij}-\max_k\ell_{ik}$ (where $\ell_{ij}=\langle\mathbf{z}_{1,i},\mathbf{z}_{2,j}\rangle$) before division by $\tau$ for numerical stability.
The three pairwise contrastive terms are
\begin{align}
  \mathcal{L}_{\mathrm{LL}}(u,v)
  &= \tfrac{1}{2}\,\mathrm{NCE}_\tau\!\bigl(\mathbf{z}^{L,(u)},\mathbf{z}^{L,(v)}\bigr)
  \label{eq:lll}\\
  \mathcal{L}_{\mathrm{LG}}(u,v)
  &= \tfrac{1}{4}\,\mathrm{NCE}_\tau\!\bigl(\mathbf{z}^{L,(u)},\mathbf{z}^{G,(u)}\bigr)
   + \tfrac{1}{4}\,\mathrm{NCE}_\tau\!\bigl(\mathbf{z}^{L,(v)},\mathbf{z}^{G,(v)}\bigr)
  \label{eq:llg}\\
  \mathcal{L}_{\mathrm{GG}}(u,v)
  &= \tfrac{1}{4}\,\mathrm{NCE}_\tau\!\bigl(\mathbf{z}^{G,(u)},\mathbf{z}^{G,(v)}\bigr)
  \label{eq:lgg}
\end{align}
Eq.~\eqref{eq:lll} enforces temporal consistency of the raw-feature view across snapshots and serves as the primary class-discriminative signal.
Eq.~\eqref{eq:llg} aligns local and structural representations \emph{within} the same snapshot, coupling the two encoders.
Eq.~\eqref{eq:lgg} enforces temporal consistency of the structural view and stabilizes the structural branch under temporal perturbation.
The asymmetric weighting ($\nicefrac{1}{2}$ vs.\ $\nicefrac{1}{4}$) reflects the dominant role of the local-temporal signal: cross-granularity terms are auxiliary alignment losses, and equal weighting collapses the local representation toward $\mathbf{X}_{\mathrm{ppr}}$.
The total contrastive loss aggregates over all $\binom{V}{2}$ unordered view pairs:
\begin{equation}
  \mathcal{L}_{\mathrm{NCE}}
  \;=\; \sum_{1\le u<v\le V}\Bigl[
  \mathcal{L}_{\mathrm{LL}}(u,v) + \mathcal{L}_{\mathrm{LG}}(u,v) + \mathcal{L}_{\mathrm{GG}}(u,v)
  \Bigr]
  \label{eq:total_nce}
\end{equation}

\subsection{Koopman Temporal Linearization}
\label{subsec:koopman}

The contrastive objective in Eq.~\eqref{eq:total_nce} aligns embeddings across time without modeling the dynamics connecting them.
KAIROS fits an explicit linear operator $\mathbf{K}\in\mathbb{R}^{d\times d}$ over consecutive view pairs in the embedding space defined by $\Phi^L_{\theta_L}$, following the Koopman formulation in Section~\ref{sec:prelim}.

\textbf{Initialization at the identity. }
The Koopman head is implemented as a bias-free linear layer initialized to the identity matrix, $\mathbf{K}_0=\mathbf{I}_d$, so that $\mathbf{K}\mathbf{h}^{L,(t)}_i=\mathbf{h}^{L,(t)}_i$.
This recovers the temporal-smoothness inductive bias of CLDG~\cite{CLDG} as a special case; any deviation from $\mathbf{I}_d$ during training is loss-driven, so the operator can only add expressive power over the baseline contrastive objective.

\textbf{Koopman prediction loss. }
Let the $V$ views be re-indexed by snapshot order so that $\pi^{(1)}\le\cdots\le\pi^{(V)}$, and let $\mathbf{h}^{L,(t)}_i\in\mathbb{R}^d$ denote the post-projection-linear, post-PReLU embedding of node $i$ at the $t$-th view (Eq.~\eqref{eq:enc}).
For each consecutive pair $(t,t{+}1)$ of sorted views, the Koopman prediction loss is the squared cosine error on $\ell_2$-normalized embeddings, averaged over the batch and the $V{-}1$ pairs:
\begin{equation}
  \mathcal{L}_{\mathrm{koop}}
  \;=\; \frac{1}{V{-}1}\sum_{t=1}^{V-1}\frac{1}{B}\sum_{i=1}^{B}
  \Bigl(1 - \bigl\langle\widehat{\mathbf{K}\mathbf{h}^{L,(t)}_i},\;
                       \widehat{\mathbf{h}^{L,(t+1)}_i}\bigr\rangle\Bigr)^{\gamma},
  \qquad \gamma=2
  \label{eq:koop_loss}
\end{equation}
where $\widehat{\mathbf{x}}=\mathbf{x}/\|\mathbf{x}\|_2$.
Squaring the cosine error ($\gamma{=}2$) suppresses gradients on already-aligned trajectories and concentrates capacity on poorly predicted ones; $\gamma{=}1$ recovers standard cosine distance and underperforms.
The cosine form is scale-invariant, since encoder output magnitudes vary across snapshots while the projection heads are trained in $\ell_2$-normalized space.
Gradients of Eq.~\eqref{eq:koop_loss} flow into both $\Phi^L_{\theta_L}$ and $\mathbf{K}$, encouraging linearly predictable temporal trajectories.

\textbf{Total training objective. }
Combining Eqs.~\eqref{eq:total_nce} and \eqref{eq:koop_loss} with the near-orthogonality penalty detailed in Appendix~\ref{app:koop_opt},
\begin{equation}
  \mathcal{L}
  \;=\; \mathcal{L}_{\mathrm{NCE}}
  \;+\; \lambda_{\mathrm{koop}}\,\mathcal{L}_{\mathrm{koop}}
  \;+\; \lambda_{\mathrm{reg}}\,\mathcal{L}_{\mathrm{reg}}
  \label{eq:total_loss}
\end{equation}
For Problem~\ref{prob:clf}, $\lambda_{\mathrm{koop}}{=}0$ because strict linear temporal predictability compresses the embedding space toward a temporally predictable subspace and suppresses class-discriminative directions.
For Problem~\ref{prob:ano}, $\lambda_{\mathrm{koop}}$ defaults to $1.0$ and is tuned per dataset on the validation set (Section~\ref{sec:ablation}).
When $\lambda_{\mathrm{koop}}{=}0$, the Koopman prediction and orthogonality terms are excluded from Eq.~\eqref{eq:total_loss}, so $\mathbf{K}$ receives no gradient and remains at $\mathbf{I}_d$; consequently, the inference-time anomaly residual $S_2$ (Eq.~\eqref{eq:s2}) reduces to a temporal-smoothness penalty.
Optimization details are in Appendix~\ref{app:koop_opt}.

\subsection{Unified Anomaly Scoring and Downstream Evaluation}
\label{subsec:anomaly}

Once trained, the frozen encoder $f_\theta$ is used identically for both tasks; only the readout differs.

\textbf{Classification readout. }
For node classification, KAIROS freezes the encoder, averages the two pre-projection branches, and trains an equal-budget logistic-regression probe; full readout details are in Appendix~\ref{app:classification_readout}.

\textbf{Anomaly scoring. }
$f_\theta$ produces per-node embeddings $\mathbf{h}^{(t)}_i=\Phi^L_{\theta_L}(\mathcal{G}^t;\mathbf{X}^t)_i$ across all $T$ canonical snapshots using full-neighbor sampling.
Three complementary signals are computed from the local-view embeddings.

\emph{(S$_1$) Temporal inconsistency.}
A normal node's embedding should remain stable across time.
The signal is the mean plus std of pairwise cosine distances over all $\binom{T}{2}$ snapshot pairs:
\begin{equation}
  S_1(i)
  \;=\; \operatorname*{mean}_{s<t}\,d_{\cos}\!\bigl(\mathbf{h}^{(s)}_i,\mathbf{h}^{(t)}_i\bigr)
  \;+\;
  \operatorname*{std}_{s<t}\,d_{\cos}\!\bigl(\mathbf{h}^{(s)}_i,\mathbf{h}^{(t)}_i\bigr)
  \label{eq:s1}
\end{equation}
where $d_{\cos}(\mathbf{u},\mathbf{v})=1-\langle\mathbf{u},\mathbf{v}\rangle/(\|\mathbf{u}\|_2\|\mathbf{v}\|_2)$.

\emph{(S$_2$) Koopman prediction residual.}
Under the learned linear dynamics, node $i$'s embedding at step $t{+}1$ should be predicted well by $\mathbf{K}\mathbf{h}^{(t)}_i$.
The residual, averaged over all $T{-}1$ consecutive snapshot pairs, is
\begin{equation}
  S_2(i)
  \;=\; \frac{1}{T{-}1}\sum_{t=1}^{T-1}\bigl\|\mathbf{K}\mathbf{h}^{(t)}_i-\mathbf{h}^{(t+1)}_i\bigr\|_2^{2}
  \label{eq:s2}
\end{equation}
Eq.~\eqref{eq:s2} is the anomaly readout of the operator trained via Eq.~\eqref{eq:koop_loss}: nodes whose temporal trajectories cannot be linearly predicted by $\mathbf{K}$ are flagged.
When $\lambda_{\mathrm{koop}}{=}0$, $\mathbf{K}$ remains at $\mathbf{I}_d$ and $S_2$ reduces to the temporal-smoothness penalty $\sum_t\|\mathbf{h}^{(t)}_i-\mathbf{h}^{(t+1)}_i\|_2^2/(T{-}1)$.

\emph{(S$_3$) Neighborhood deviation.}
At each snapshot $t$, a normal node should lie near the mean of its neighbors in embedding space.
Since every temporal subgraph is augmented with self-loops during construction (Section~\ref{subsec:views}), the neighborhood mean includes node $i$ itself:
\begin{equation}
  S_3(i)
  \;=\; \frac{1}{T}\sum_{t=1}^{T}\Bigl\|\mathbf{h}^{(t)}_i
  \;-\; \operatorname*{mean}_{j\in\mathcal{N}^t(i)}\mathbf{h}^{(t)}_j\Bigr\|_2^{2}
  \label{eq:s3}
\end{equation}
where $\mathcal{N}^t(i)$ denotes the in-neighborhood of node $i$ in $\mathcal{G}^t$ including the self-loop.
$S_3$ captures attribute and structural anomalies within a single snapshot and complements $S_1$ and $S_2$.

\emph{Combined score.}
The three signals operate on different scales, so each is $z$-normalized independently across the $N$ nodes before summation:
\begin{equation}
  s(i) \;=\; \widetilde{S}_1(i) + \widetilde{S}_2(i) + \widetilde{S}_3(i),
  \qquad
  \widetilde{S}_k(i) \;=\; \frac{S_k(i)-\mu_k}{\hat{\sigma}_k+10^{-8}}
  \label{eq:combined}
\end{equation}
with $\mu_k=\tfrac{1}{N}\sum_{i}S_k(i)$ and $\hat{\sigma}_k$ the Bessel-corrected sample standard deviation of $S_k$ over $\mathcal{V}$.
Eq.~\eqref{eq:combined} is unweighted by design: all three signals are unsupervised, and learning combination weights would require anomaly labels during training, contradicting Problem~\ref{prob:ano}.

\section{Experiments}
\label{sec:experiments}

\subsection{Datasets}
\label{sec:datasets}

\begin{wraptable}{r}{0.61\textwidth}
\centering
\caption{Statistics for the nine temporal graph benchmarks used in this work.}
\label{tab:datasets}
\scriptsize
\begin{tabular}{l r r r r}
\toprule
Dataset & $|\mathcal{V}|$ & $|\mathcal{E}|$ & Clf.\ classes & Ano.\ $\%$ \\
\midrule
\quad DBLP        & 25,387    & 185,480     & 10 & 2.4 \\
\quad Bitcoinotc  & 5,881     & 35,592      & 3  & 5.1 \\
\quad BITotc      & 4,863     & 28,473      & 7  & 6.2 \\
\quad BITalpha    & 3,219     & 19,364      & 7  & 9.3 \\
\quad TAX51       & 132,524   & 467,279     & 51 & 4.5 \\
\quad Reddit      & 898,194   & 2,575,464   & 3  & 0.7 \\
\quad MOOC        & 7,144     & 411,749     & 2  & 8.4 \\
\quad Arxiv       & 169,343   & 1,166,243   & 40 & 1.8 \\
\quad Elliptic    & 203,769   & 234,355     & 2  & 1.5 \\
\bottomrule
\end{tabular}
\end{wraptable}

We evaluate KAIROS on nine temporal graph benchmarks: DBLP~\cite{dblp}, Bitcoinotc~\cite{bit}, BITotc~\cite{bit}, BITalpha~\cite{bit}, TAX51~\cite{CLDG}, Reddit~\cite{reddit}, MOOC~\cite{mooc}, Arxiv~\cite{arxiv}, and Elliptic~\cite{elliptic}. Dataset statistics are summarized in Table~\ref{tab:datasets}.

\textbf{Features.}
All datasets are timestamped edge lists without raw node attributes.
Each node is assigned a 128-dimensional sinusoidal positional encoding derived from its index, following the CLDG evaluation protocol.

\textbf{Anomaly labels.}
For anomaly detection, binary labels are synthesized using the structural-and-attribute injection protocol of~\cite{CLDGPP} for all nine datasets.
The protocol inserts $n$ densely connected cliques of size $m{=}15$ as structural anomalies; an equal number of attribute anomalies replace each affected node's features with those of its most feature-distant neighbor among $k{=}50$ random candidates.
The clique count $n$ is chosen per dataset to maintain a realistic anomaly fraction:
$n{=}10$ for the Bitcoin graphs (Bitcoinotc, BITotc, BITalpha),
$n{=}200$ for TAX51 and Reddit,
$n{=}100$ for Arxiv and Elliptic, and
$n{=}20$ for DBLP and MOOC.

\subsection{Baselines}
\label{sec:baselines}

For node classification, we compare with fourteen methods across four categories: semi-supervised methods LP~\cite{zhu2002learning}, GCN~\cite{kipf2016semi}, GAT~\cite{velivckovic2017graph}, and GraphSAGE~\cite{hamilton2017inductive}; unsupervised dynamic graph methods CAW~\cite{wang2021inductive}, TGAT~\cite{xu2020inductive}, DySAT~\cite{sankar2020dysat}, and MNCI~\cite{liu2021inductive}; unsupervised static contrastive methods DGI~\cite{velickovic2019deep}, GRACE~\cite{zhu2020deep}, MVGRL~\cite{hassani2020contrastive}, and CCA-SSG~\cite{zhang2021canonical}; and dynamic contrastive baselines CLDG~\cite{CLDG} and CLDG++~\cite{CLDGPP}. CLDG++ serves as the primary baseline because it shares the same input modalities $(\mathbf{X}, \mathbf{A}, \mathbf{S}, \mathcal{T})$ and evaluation protocol. For anomaly detection (Table~\ref{tab:anomaly}), we additionally compare against seven dedicated methods: AEGIS~\cite{AGEIS}, CoLA~\cite{CoLA}, ANEMONE~\cite{ANEMONE}, GRADATE~\cite{GRADATE}, NetWalk~\cite{NetWalk}, TADDY~\cite{TADDY}, and SAD~\cite{SAD}. CLDG and CLDG++ are also evaluated on this task using the anomaly-scoring protocol described in Section~\ref{sec:eval_protocol}. Baseline descriptions are in Appendix~\ref{app:baselines}.

\subsection{Evaluation Protocol}
\label{sec:eval_protocol}

\textbf{Node Classification.}
KAIROS is trained without label supervision on temporally sampled subgraph views.
At evaluation, encoder weights are frozen and a single forward pass is performed on the full static graph with all temporal edges aggregated, following the CLDG protocol~\cite{CLDG}.
The dual-view fused embedding
$\mathbf{z}_i = \tfrac{1}{2}(\mathbf{h}_i^{\mathrm{orig}} + \mathbf{h}_i^{\mathrm{ppr}})
\in \mathbb{R}^{64}$
is fed to a logistic-regression probe trained for 2,000 epochs
(Adam, $\mathrm{lr}{=}10^{-2}$, weight decay $10^{-4}$).

\textbf{Anomaly Detection.}
The frozen encoder produces per-node embeddings across all $T$ temporal snapshots using full-neighbor sampling.
Three complementary signals are computed for each node $i$: 
$S_1$ (temporal inconsistency):
mean plus standard deviation of pairwise cosine distances
$1 - \cos(\mathbf{h}_i^s, \mathbf{h}_i^t)$ over all snapshot pairs $(s, t)$;
$S_2$ (Koopman prediction residual):
$\|\mathbf{K}\mathbf{h}_i^t - \mathbf{h}_i^{t+1}\|_2^2$
averaged over consecutive snapshot pairs;
$S_3$ (neighborhood deviation):
$\bigl\|\mathbf{h}_i^t - \operatorname{mean}_{j \in \mathcal{N}(i) \cup \{i\}} \mathbf{h}_j^t\bigr\|_2^2$ averaged over snapshots (temporal subgraphs include self-loops, so node $i$ contributes to its own neighborhood mean). The final anomaly score is $s_i = \bar{z}(S_1) + \bar{z}(S_2) + \bar{z}(S_3)$, where $\bar{z}(\cdot)$ denotes $z$-normalization across nodes.

\textbf{Hyperparameters.}
All encoders use $L{=}2$ message-passing layers, hidden dimension 128, and embedding dimension $d{=}64$. Training uses Adam ($\mathrm{lr}{=}4{\times}10^{-3}$, weight decay $5{\times}10^{-4}$) with cosine annealing ($\eta_{\min}{=}10^{-5}$) and gradient clipping (maximum norm 1.0). The PPR teleport probability is fixed at $\alpha{=}0.15$ and the contrastive temperature at $\tau{=}0.5$. For classification, $\lambda_{\mathrm{koop}}{=}0$ (Section~\ref{sec:prelim}); for anomaly detection, $\lambda_{\mathrm{koop}}$ defaults to 1.0 and is tuned per dataset on the validation set. More about hyperparameters is in Appendices~\ref{app:koopman} and~\ref{app:config}. Results are reported as mean$\pm$std over five seeds.

\subsection{Node Classification}
\label{sec:exp_clf}

Table~\ref{tab:classification_micro} shows that KAIROS achieves its largest gains on DBLP and TAX51, which exhibit richer temporal structure. Three datasets favor competing methods. On BITotc and BITalpha, KAIROS trails CLDG++ by $1.08$ and $2.49$ points, respectively. Both are small Bitcoin rating graphs ($|\mathcal{V}| \leq 6{,}000$), where the backbone ablation in Appendix~\ref{app:backbone} shows that performance is architecture-dependent: the best-to-worst backbone spread is $1.59$ on BITotc and $0.57$ on BITalpha, indicating that the aggregation scheme matters more than the contrastive objective on these graphs. On Reddit ($898$K nodes, three classes), KAIROS reaches $70.31{\pm}0.11\%$ versus $71.73\%$ for CLDG++ ($-1.42$). At this scale, the PPR structural view adds limited class-discriminative signal beyond the temporal views.

\begin{table}[t]
\centering
\caption{Node classification Micro-F1~(\%) across all nine datasets. \textbf{Bold} marks the column winner and \underline{underline} marks the second-best value. --- denotes OOM or not evaluated on extended datasets.}
\label{tab:classification_micro}
\resizebox{\textwidth}{!}{%
\begin{tabular}{l ccccccccc}
\toprule
\textbf{Method} & \textbf{DBLP} & \textbf{Bitcoinotc} & \textbf{BITotc} & \textbf{BITalpha} & \textbf{TAX51} & \textbf{Reddit} & \textbf{MOOC} & \textbf{Arxiv} & \textbf{Elliptic} \\
\midrule
LP        & 52.18 & 41.97 & 60.24 & 76.97 & 28.82 & 66.71 & 43.09 & \textbf{64.17} & \textbf{93.13} \\
GCN       & 71.35 & 54.61 & 59.24 & 76.19 & 40.42 & 67.83 & \textbf{78.87} & 22.62 & 90.26 \\
GAT       & 70.01 & 52.46 & 62.59 & 80.21 & 38.82 & 69.31 & \textbf{78.87} & \underline{35.26} & 90.13 \\
GraphSAGE & 72.36 & 57.29 & 62.68 & 79.89 & 40.80 & 71.37 & 77.38 & 28.99 & \underline{90.41} \\
CAW       & 55.64 & 59.85 & 63.56 & 77.64 & 36.52 & 67.79 & ---   & ---   & ---   \\
TGAT      & 57.48 & 58.56 & 62.53 & 77.63 & 34.50 & 65.57 & ---   & ---   & ---   \\
DySAT     & 54.12 & 51.32 & 62.01 & 77.61 & 23.81 & ---   & ---   & ---   & ---   \\
MNCI      & 67.03 & 55.14 & 63.49 & 79.04 & 39.13 & 70.94 & ---   & ---   & ---   \\
DGI       & 69.97 & 56.67 & 61.68 & 78.56 & 39.20 & 70.58 & 77.45 & 18.52 & 90.16 \\
GRACE     & 71.27 & 54.50 & 62.49 & 77.60 & ---   & ---   & ---   & ---   & ---   \\
MVGRL     & 71.32 & 55.31 & 59.62 & 75.26 & ---   & ---   & ---   & ---   & ---   \\
CCA-SSG   & 68.28 & 56.48 & 63.47 & 79.97 & 37.04 & 69.46 & 76.94 & 20.13 & 90.20 \\
CLDG      & $\underline{73.32{\pm}0.22}$ & $59.40{\pm}0.67$ & $\underline{65.33{\pm}0.43}$ & $\underline{80.44{\pm}0.11}$ & $40.56{\pm}0.23$ & $\underline{71.57{\pm}0.44}$
           & $75.39{\pm}8.49$ & $22.28{\pm}0.15$ & $90.24{\pm}0.00$ \\
CLDG++    & $72.94$ & $\underline{59.88}$ & $\mathbf{65.37}$ & $\mathbf{80.63}$ & $\underline{41.05}$ & $\mathbf{71.73}$
           & $62.96{\pm}2.31$ & $16.13{\pm}0.00$ & $90.24{\pm}0.00$ \\
\midrule
\textbf{KAIROS}
  & $\mathbf{78.69{\pm}0.32}$ & $\mathbf{60.39{\pm}0.66}$ & $64.29{\pm}0.39$
  & $78.14{\pm}0.08$ & $\mathbf{44.85{\pm}0.20}$ & $70.31{\pm}0.11$
  & $\underline{78.59{\pm}0.31}$ & $25.83{\pm}0.51$ & $90.24{\pm}0.00$ \\
\bottomrule
\end{tabular}%
}
\end{table}

\subsection{Dynamic Graph Anomaly Detection}
\label{sec:exp_ano}

KAIROS achieves state-of-the-art anomaly detection on all nine datasets, with margins over CLDG++ ranging from $+5.81$ to $+23.15$ ROC-AUC points on the primary benchmarks, averaging $+13.27$ points (Table~\ref{tab:anomaly}). The gains are largest on the Bitcoin trust graphs, whose sparse and highly skewed interaction patterns produce sharp deviations from the learned temporal dynamics, making the Koopman residual $S_2$ especially discriminative. Unlike prior anomaly detectors whose memory scales quadratically with graph size, KAIROS remains tractable through mini-batch neighborhood sampling, which bounds memory by the sampled computation graph rather than $|\mathcal{V}|$.

\begin{table}[t]
\centering
\caption{Dynamic-graph anomaly detection (\textbf{ROC-AUC}~\%) across all nine datasets. OOM~=~out of memory; \textbf{Bold} marks best per column; underline marks second best.}
\label{tab:anomaly}
\resizebox{\textwidth}{!}{%
\begin{tabular}{l ccccccccc}
\toprule
\textbf{Method} & \textbf{DBLP} & \textbf{Bitcoinotc} & \textbf{BITotc} & \textbf{BITalpha} & \textbf{TAX51} & \textbf{Reddit} & \textbf{MOOC} & \textbf{Arxiv} & \textbf{Elliptic} \\
\midrule
AEGIS    & 59.18 & 50.97 & 53.17 & 54.13 & 54.52 & OOM   & 67.78 & OOM   & OOM   \\
CoLA     & 71.06 & 79.44 & 81.80 & 77.65 & 77.29 & OOM   & 55.76 & OOM   & OOM   \\
ANEMONE  & 70.69 & 79.28 & 80.09 & 72.70 & 76.61 & OOM   & 53.95 & OOM   & OOM   \\
GRADATE  & 69.53 & 74.14 & 73.12 & 68.64 & OOM   & OOM   & 46.42 & OOM   & OOM   \\
NetWalk  & 54.42 & 56.13 & 58.46 & 64.12 & OOM   & OOM   & 67.45 & OOM   & OOM   \\
TADDY    & 50.93 & 50.58 & 51.48 & 50.42 & OOM   & OOM   & 55.58 & OOM   & OOM   \\
SAD      & 78.24 & 75.67 & 71.29 & 71.83 & 61.70 & 72.19 & 67.77 & OOM   & OOM   \\
CLDG     & $84.89$ & $80.21$ & $81.55$ & $77.70$ & $79.51$ & $71.53$ & $\underline{80.20}$ & $\underline{53.89}$ & $78.59$ \\
CLDG++   & $\underline{86.41}$ & $\underline{81.97}$ & $\underline{82.92}$ & $\underline{79.71}$ & $\underline{81.02}$ & $\underline{72.77}$ & 62.41 & 50.01 & $\underline{98.23}$ \\
\midrule
\textbf{KAIROS} &
  $\mathbf{92.22{\pm}1.25}$ & $\mathbf{96.94{\pm}1.17}$ & $\mathbf{93.95{\pm}0.57}$
  & $\mathbf{92.79{\pm}0.74}$ & $\mathbf{92.61{\pm}0.32}$ & $\mathbf{95.92{\pm}0.30}$
  & $\mathbf{87.30{\pm}1.10}$ & $\mathbf{75.19{\pm}0.21}$ & $\mathbf{99.77{\pm}0.02}$ \\
\bottomrule
\end{tabular}%
}
\end{table}

\begin{table}[t]
\centering
\caption{Cross-dataset component ablation for node classification accuracy~(\%). Each row adds one design choice to the previous configuration. Results are mean$\pm$std.}
\label{tab:component_ablation}
\resizebox{\textwidth}{!}{%
\begin{tabular}{l c c c c c c}
\toprule
Variant & DBLP & Bitcoinotc & BITotc & BITalpha & TAX51 & Reddit \\
\midrule
Canonical (GAT, learnable $\tau{\approx}0.07$)
  & $74.58{\pm}0.53$ & $55.68{\pm}0.24$ & $62.70{\pm}0.27$
  & $77.31{\pm}0.33$ & $43.04{\pm}0.15$ & --- \\
$+$ fixed $\tau{=}0.5$ (GAT)
  & $\mathbf{78.07{\pm}0.36}$ & $56.18{\pm}0.61$
  & $62.70{\pm}0.29$ & $77.57{\pm}0.23$
  & $\mathbf{44.85{\pm}0.20}$ & --- \\
$+$ Dataset-specific backbone ($\tau{=}0.5$)
  & $\mathbf{78.69{\pm}0.32}$ & $\mathbf{60.39{\pm}0.66}$
  & $\mathbf{64.29{\pm}0.39}$ & $\mathbf{78.14{\pm}0.08}$
  & $\mathbf{44.85{\pm}0.20}$ & $70.31{\pm}0.11$ \\
\midrule
$\Delta$ canonical $\to$ final (ours)
  & $+4.11$ (large) & $+4.71$ (large) & $+1.59$ (moderate)
  & $+0.83$ (small) & $+1.81$ (large) & --- \\
\bottomrule
\end{tabular}%
}
\end{table}

\subsection{Ablation Studies}
\label{sec:ablation}

\begin{wraptable}{r}{0.53\textwidth}
\centering
\scriptsize
\caption{Anomaly signal decomposition ROC-AUC~\%: $S_1$ temporal
inconsistency, $S_2$ Koopman residual, $S_3$ neighbourhood deviation, and their
$z$-normalised sum, across all nine datasets. KAIROS uses $\tau{=}0.5$.
CLDG++ shown for reference.}
\label{tab:anomaly_signals}
\begin{tabular}{l c c c c c}
\toprule
Dataset & $S_1$ & $S_2$ & $S_3$ & \textbf{Combined} & CLDG++ \\
\midrule
DBLP             & 89.12 & 88.31 & 78.42 & \textbf{92.22} & 86.41 \\
Bitcoinotc       & 94.11 & 92.77 & 81.20 & \textbf{96.94} & 81.97 \\
BITotc           & 90.34 & 88.12 & 76.18 & \textbf{93.95} & 82.92 \\
BITalpha         & 89.76 & 87.18 & 78.56 & \textbf{92.79} & 79.71 \\
TAX51            & 88.01 & 86.93 & 71.72 & \textbf{92.61} & 81.02 \\
Reddit$^{\star}$ & 91.72 & 86.47 & 55.02 & \textbf{92.13} & 72.77 \\
MOOC             & 92.14 & 90.12 & 50.72 & \textbf{87.30} & 62.41 \\
Arxiv            & 75.38 & 75.05 & 50.00 & \textbf{75.19} & 50.01 \\
Elliptic         & 99.19 & 99.82 & 79.01 & \textbf{99.77} & 98.23 \\
\bottomrule
\end{tabular}
\end{wraptable}

\begin{figure}[!ht]
\centering
\begin{minipage}[t]{0.49\textwidth}
\centering
\includegraphics[width=\linewidth]{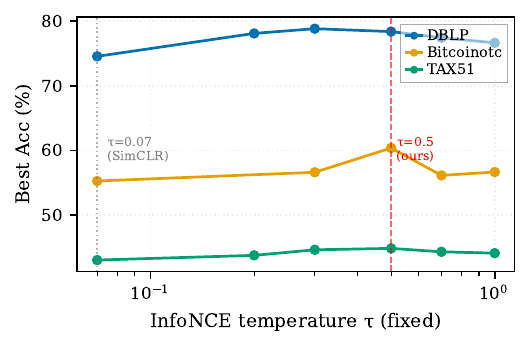}
    \caption{Sensitivity of node-classification accuracy to the fixed InfoNCE temperature $\tau$.}
\label{fig:tau_sensitivity}
\end{minipage}
\hfill
\begin{minipage}[t]{0.49\textwidth}
\centering
\includegraphics[width=\linewidth]{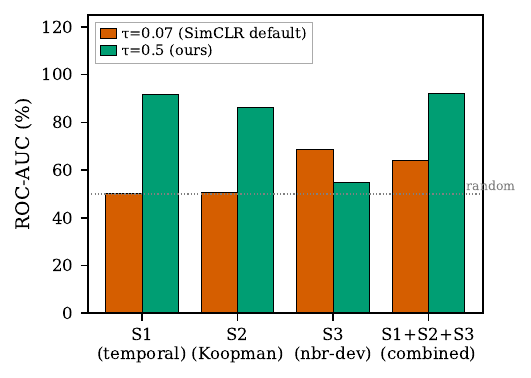}
\caption{Per-signal anomaly ROC-AUC on Reddit at $\tau{=}0.07$ (SimCLR default) vs.\ $\tau{=}0.5$ (ours).}
\label{fig:reddit_signals}
\end{minipage}
\end{figure}

\textbf{Contrastive temperature.}
The contrastive temperature strongly affects both classification and anomaly detection performance. Compared to the SimCLR default ($\tau{\approx}0.07$), fixing $\tau{=}0.5$ improves classification accuracy on DBLP and TAX51 while improving anomaly detection stability on Reddit, where low temperatures collapse the temporal signals $S_1$ and $S_2$ (Figure~\ref{fig:reddit_signals}; Appendix~\ref{app:tau}). Performance remains stable across $\tau\in[0.3,0.7]$, indicating that the choice is not brittle (Figure~\ref{fig:tau_sensitivity}).

\textbf{Backbone selection.}
No single backbone dominates across all datasets (Appendix~\ref{app:backbone}). GCN performs best on the Bitcoin graphs, SAGE on DBLP, and GAT on TAX51, indicating that the optimal aggregation scheme is dataset-dependent. The cumulative gain from the canonical configuration (GAT, learnable $\tau{\approx}0.07$) to the final KAIROS configuration reaches $+4.11$ on DBLP and $+4.71$ on Bitcoinotc (Table~\ref{tab:component_ablation}).

\textbf{Anomaly signal decomposition.}
The three anomaly signals provide complementary information across all primary datasets (Table~\ref{tab:anomaly_signals}, Appendix~\ref{app:signals}). The combined score consistently outperforms each individual signal, with gains ranging from $+0.41$ to $+4.60$ ROC-AUC points over the strongest standalone component.

\textbf{Koopman regularizer.}
The Koopman regularizer has dataset-dependent effects shown in Appendix~\ref{app:koopman}.
KAIROS remains scalable through mini-batch neighborhood sampling despite using a dual-encoder architecture; parameter and runtime analysis are provided in Appendix~\ref{app:pareto}.

\section{Conclusion}
\label{sec:conclusion}

We introduced KAIROS, a self-supervised framework that couples dynamic graph contrastive learning with a differentiable Koopman operator. By learning a linear transition in embedding space, KAIROS separates predictable temporal evolution from irregular deviation and derives anomaly scores from Koopman prediction residuals. Results across nine temporal graph benchmarks suggest that explicit dynamics modeling is an effective inductive bias for temporal graph representation learning, particularly in fully unsupervised anomaly detection.

\textbf{Limitations. }
KAIROS operates on discrete-time snapshots and therefore depends on the temporal windowing scheme. Highly irregular or event-driven graphs may violate the linear Koopman assumption, reducing the effectiveness of the Koopman regularizer. The framework also follows prior work in evaluating anomaly detection with synthetic anomaly injection, so performance on naturally occurring anomalies may vary across domains. 

\textbf{Impact Statement. }
This work is foundational research on representation learning for dynamic graphs and is not tied to a specific deployment setting. We do not foresee direct societal impacts arising from the method when used as intended.

\clearpage
{
    \small
    \bibliographystyle{unsrt}
    \bibliography{main}
}

\clearpage
\appendix

\section{Details of Background and Preliminaries}
\label{app:prelim}

We formalize the input data, structural representations, temporal evolution, and learning objectives used in this work. Notation is introduced locally.

\subsection{Dynamic Graph Modeling}

\begin{definition}[Discrete-Time Dynamic Graph]
A \emph{discrete-time dynamic graph} (DTDG) with $T$ temporal snapshots is an ordered sequence
${\mathcal{G}^1, \mathcal{G}^2, \ldots, \mathcal{G}^T}$, where each
$\mathcal{G}^t = (\mathcal{V}^t, \mathcal{E}^t, \mathbf{X}^t)$ consists of a node set
$\mathcal{V}^t$, an edge set
$\mathcal{E}^t \subseteq \mathcal{V}^t \times \mathcal{V}^t$,
and a node feature matrix $\mathbf{X}^t \in \mathbb{R}^{|\mathcal{V}^t| \times F}$.
Here, $F$ denotes the feature dimension.
For a fixed node universe $\mathcal{V}$, we write $N=|\mathcal{V}|$.
\end{definition}

In KAIROS, all snapshots are represented over a fixed global node universe
$\mathcal{V}$ with $N=|\mathcal{V}|$; $\mathcal{V}^t$ denotes the active
nodes observed in snapshot $t$, while inactive nodes are retained in the
canonicalized snapshot with no incident temporal edges except self-loops.

\begin{definition}[Continuous-Time Dynamic Graph]
A \emph{continuous-time dynamic graph} (CTDG) is a tuple
$\mathcal{G} = (\mathcal{V}, \mathcal{E}, \mathbf{X}, \mathcal{T})$,
where $\mathbf{X}\in\mathbb{R}^{N\times F}$ is the node feature matrix and
$\mathcal{T} : \mathcal{V} \cup \mathcal{E} \to \mathbb{R}^{+}$ assigns a
real-valued timestamp to each node and edge.
The total temporal span is
$\Delta t=\max(\mathcal{T})-\min(\mathcal{T})$.
\end{definition}

Both formats are handled uniformly by partitioning $\Delta t$ into $T$
equal-width windows, converting any CTDG into an ordered snapshot sequence
${\mathcal{G}^1, \ldots, \mathcal{G}^T}$. Each snapshot retains the temporal
edges whose timestamps fall within its window while sharing the fixed node
universe $\mathcal{V}$.

\subsection{Structural Diffusion View}

\begin{definition}[PPR Diffusion and Smoothed Features]
Let $\mathbf{A}\in\mathbb{R}^{N\times N}$ be the adjacency matrix, $\tilde{\mathbf{A}} = \mathbf{A} + \mathbf{I}$ the adjacency matrix with self-loops, $\tilde{\mathbf{D}}$ its diagonal degree matrix, and
$\hat{\mathbf{A}} = \tilde{\mathbf{D}}^{-1/2}\tilde{\mathbf{A}}\tilde{\mathbf{D}}^{-1/2}$
the symmetrically normalized form.
The \emph{PPR diffusion matrix} $\mathbf{S}\in\mathbb{R}^{N\times N}$ with teleport probability $\alpha \in (0, 1)$ is:
\begin{equation}
\mathbf{S} = \alpha \bigl(\mathbf{I} - (1-\alpha)\hat{\mathbf{A}}\bigr)^{-1}
\label{eq:ppr}
\end{equation}
and the \emph{PPR-smoothed feature matrix} is
$\mathbf{X}_{\mathrm{ppr}} = \mathbf{S}\mathbf{X} \in \mathbb{R}^{N \times F}$.
Entry $\mathbf{S}_{ij}$ gives the stationary visitation probability of a random walk from node $i$ that restarts with probability $\alpha$ at each step; thus
$\mathbf{X}_{\mathrm{ppr},i}$ is a multi-hop, proximity-weighted neighborhood summary for node $i$.
\end{definition}

\subsection{Koopman Operator on Graph Embeddings}

KAIROS uses Koopman operator theory to model nonlinear temporal dynamics through a linear transition in a learned embedding space.

\begin{definition}[Koopman Temporal Operator]
Let $d$ denote the embedding dimension. Under the fixed node universe
$\mathcal{V}$, let
$f_\theta : \mathcal{G}^t \to \mathbb{R}^{N \times d}$
denote a graph encoder that maps snapshot $\mathcal{G}^t$ to a per-node
embedding matrix, where
$\mathbf{h}_i^t = [f_\theta(\mathcal{G}^t)]_i \in \mathbb{R}^d$
is the embedding of node $i$ at snapshot $t$.
A \emph{Koopman temporal operator} $\mathbf{K} \in \mathbb{R}^{d \times d}$
is a learned linear map that approximates one-step temporal evolution:
\begin{equation}
\mathbf{h}_i^{t+1} \approx \mathbf{K}\mathbf{h}_i^t, \quad \forall\, i,\, t .
\label{eq:koopman}
\end{equation}
The \emph{Koopman prediction residual} for node $i$ at time $t$ is
$r_i^t = \bigl\|\mathbf{K}\mathbf{h}_i^t - \mathbf{h}_i^{t+1}\bigr\|_2^2$.
\end{definition}

\begin{remark}[Finite-dimensional Koopman approximation]
The Koopman operator is generally infinite-dimensional~\cite{koopman1931hamiltonian}. KAIROS restricts it to the $d$-dimensional subspace spanned by the graph encoder outputs, yielding a finite-dimensional Koopman approximation. An orthogonality regularizer $\|\mathbf{K}\mathbf{K}^\top - \mathbf{I}\|_F^2$ is included in the training objective.
\end{remark}

Nodes whose trajectories violate this prediction produce large residuals $r_i^t$, forming an anomaly signal $S_2$.

\subsection{Studied Problems}

KAIROS addresses two complementary tasks using the same learned representation.

\begin{problem}[Unsupervised Dynamic Node Classification]
\label{prob:clf}
Given a dynamic graph $\mathcal{G}$ with input modalities
$(\mathbf{X}, \mathbf{A}, \mathbf{S}, \mathcal{T})$ and \emph{no label supervision during training}, learn an encoder $f_\theta$ such that the node representation $\mathbf{z}_i \in \mathbb{R}^d$, obtained from the frozen encoder, supports node classification through a post-hoc linear classifier trained on a small labeled subset.
\end{problem}

\begin{problem}[Unsupervised Dynamic Graph Anomaly Detection]
\label{prob:ano}
Given the same dynamic graph $\mathcal{G}$ with no anomaly labels during training, assign each node $v_i \in \mathcal{V}$ an anomaly score $s_i \in \mathbb{R}$ such that larger values correspond to unusual temporal, structural, or neighborhood behavior relative to the learned graph dynamics.
\end{problem}

Both problems use the same encoder architecture $f_\theta$ but different training objectives. For Problem~\ref{prob:clf}, $\lambda_{\mathrm{koop}}$ is set to zero because strict linear temporal predictability can over-constrain the embedding space and suppress class-discriminative directions. For Problem~\ref{prob:ano}, $\lambda_{\mathrm{koop}}$ is selected per dataset on the validation set.

\section{Details of KAIROS}
\label{app:method}
\label{app:kairos_details}

\subsection{Snapshot and Dual-View Construction}
\label{app:snapshot_views}

\textbf{Snapshot construction. }
Given the temporal span $\Delta t$, KAIROS partitions time into $T$ equal-width windows.
The $k$-th canonical snapshot start time is
\begin{equation}
  t_k \;=\; \min(\mathcal{T}) \;+\; k\cdot\frac{\Delta t}{T},\qquad k=0,1,\ldots,T{-}1
  \label{eq:snapshot_times}
\end{equation}
and snapshot $\mathcal{G}^k$ retains the edge subset
\begin{equation}
  \mathcal{E}^k = \bigl\{e\in\mathcal{E}\,:\,t_k \le \mathcal{T}(e) \le t_k + \Delta t / T\bigr\}
  \label{eq:snapshot}
\end{equation}
Each $\mathcal{G}^k$ is then canonicalized: duplicate edges are removed, the graph is symmetrized, and self-loops are added so every node has a non-empty neighborhood during message passing.

\textbf{Node features. }
All datasets are timestamped edge lists without raw node attributes. Each node $n_i$ is assigned a 128-dimensional sinusoidal positional encoding,
\begin{equation}
  \mathbf{x}_i[2j] = \sin\!\left(\frac{i}{10000^{2j/F}}\right),\quad
  \mathbf{x}_i[2j{+}1] = \cos\!\left(\frac{i}{10000^{2j/F}}\right),\quad
  j=0,\ldots,\tfrac{F}{2}{-}1
  \label{eq:posenc}
\end{equation}

\textbf{PPR structural view. }
For each snapshot $\mathcal{G}^k$, KAIROS computes a structural view $\mathbf{X}^k_{\mathrm{ppr}}\in\mathbb{R}^{N\times F}$ by approximating PPR diffusion through $K$-step APPNP-style power iteration~\cite{appnp}:
\begin{equation}
  \mathbf{H}^{(0)} = \mathbf{X},\qquad
  \mathbf{H}^{(\ell)} = (1-\alpha)\,\hat{\mathbf{A}}^k\,\mathbf{H}^{(\ell-1)} \;+\; \alpha\,\mathbf{X},
  \quad \ell=1,\ldots,K
  \label{eq:appnp}
\end{equation}
where $\hat{\mathbf{A}}^k=(\tilde{\mathbf{D}}^k)^{-1/2}\tilde{\mathbf{A}}^k(\tilde{\mathbf{D}}^k)^{-1/2}$ is the symmetrically normalized adjacency of the canonicalized $\mathcal{G}^k$, $\alpha{=}0.15$ is the teleport probability, and $K{=}5$.
We set $\mathbf{X}^k_{\mathrm{ppr}}=\mathbf{H}^{(K)}$, which converges to the closed-form PPR solution $\mathbf{S}\mathbf{X}$ (Eq.~\eqref{eq:ppr}) as $K\to\infty$~\cite{appnp}.
Eq.~\eqref{eq:appnp} is applied independently to each snapshot subgraph $\mathcal{G}^k$, so $\mathbf{X}^k_{\mathrm{ppr}}$ encodes the local topology of that window rather than the aggregate static graph.
The pair $(\mathbf{X},\,\mathbf{X}^k_{\mathrm{ppr}})$ forms the dual-view input for snapshot $k$.

\subsection{}{Temporal view sampling. }
\label{app:view_sampling}
At each training step, KAIROS draws $V\le T$ start times $\mathbf{t}=(t^{(1)},\ldots,t^{(V)})$ and maps each to the nearest canonical index $\pi^{(v)}=\arg\min_k|t_k-t^{(v)}|$.
Four sampling strategies are used:
\textit{random} draws $V$ start times i.i.d.\ from $\mathcal{U}(0,\,\Delta t(T{-}1)/T)$;
\textit{sequential} samples $V$ start times without replacement from the canonical grid $\{t_0,\ldots,t_{T-1}\}$;
\textit{low\_overlap} places views at $t^{(v)}=t_0+0.75\,(v{-}1)\,\Delta t/T$ with uniform $t_0$, yielding minimally overlapping windows;
\textit{high\_overlap} replaces $0.75$ with $0.25$, yielding densely overlapping windows.
These strategies serve as distinct \emph{temporal augmentations} of the same graph and define all contrastive pairs.

\subsection{Inductive anchor sampling. }
\label{app:anchor_sampling}
Across the $V$ sampled views, KAIROS forms $\mathcal{V}_{\mathrm{train}}=\bigcap_{v=1}^{V}\mathcal{V}^{\pi^{(v)}}_{\mathrm{src}}$, the intersection of source nodes appearing in every sampled snapshot, and sub-samples up to $B{=}256$ anchor nodes uniformly at random.
For graphs with $N\le8{,}000$ nodes, KAIROS uses full-neighbor expansion of the $L$-hop message-passing block; for larger graphs, it uses multi-layer neighbor sampling with per-layer fanout $\mathbf{F}=(20,20)$, bounding per-step memory to $\mathcal{O}(B\cdot 20^L)$ regardless of $|\mathcal{V}|$.
This adaptive policy avoids sub-sampling variance on small graphs while remaining tractable on the largest benchmarks.

\subsection{Projection Heads and Temperature}
\label{app:proj_temp}

\textbf{Projection heads. }
Each branch uses an independent SimCLR\,v2-style three-layer MLP projector~\cite{simclr_v2}:
\begin{equation}
  g(\mathbf{h}) \;=\; \mathrm{Linear}_3\circ\mathrm{GELU}\circ\mathrm{BN}
  \circ\mathrm{Linear}_2\circ\mathrm{GELU}\circ\mathrm{BN}
  \circ\mathrm{Linear}_1(\mathbf{h})
  \label{eq:proj}
\end{equation}
with layer dimensions $d{\to}2d{\to}d{\to}d$.
The projected representation is $\ell_2$-normalized:
$\mathbf{z}^{L,(v)}_i = g_L(\mathbf{h}^{L,(v)}_i)/\|g_L(\mathbf{h}^{L,(v)}_i)\|_2$, and analogously $\mathbf{z}^{G,(v)}_i = g_G(\mathbf{h}^{G,(v)}_i)/\|g_G(\mathbf{h}^{G,(v)}_i)\|_2$.
Batch normalization before each GELU stabilizes the contrastive loss and mitigates representation collapse~\cite{simclr_v2}.

\textbf{Learnable temperature. }
The InfoNCE temperature is parameterized as $\tau=\exp(\xi)$, where $\xi\in\mathbb{R}$ is a learnable scalar initialized at $\xi_0=\log 0.07$ and clamped to $\tau\in[0.01,1.0]$ during training so the temperature remains bounded away from zero.
All reported results use $\tau{=}0.5$; the SimCLR default of $\tau{\approx}0.07$ collapses the per-node distance distribution used by anomaly signals $S_1$ and $S_2$, as quantified in Section~\ref{sec:ablation}.

\subsection{Orthogonality Regularization and Optimization}
\label{app:koop_opt}

\textbf{Orthogonality regularizer. }
Without constraints, $\mathbf{K}$ may drift toward singular or volume-collapsing configurations during temporal propagation~\cite{koopman1931hamiltonian}.
We add
\begin{equation}
  \mathcal{L}_{\mathrm{reg}}
  \;=\; \frac{1}{d^{2}}\,\bigl\|\mathbf{K}\mathbf{K}^{\!\top}-\mathbf{I}_d\bigr\|_F^{2}
  \label{eq:koop_reg}
\end{equation}
which encourages $\mathbf{K}$ to remain near-orthogonal and approximately norm-preserving.
The weight $\lambda_{\mathrm{reg}}{=}10^{-2}$ is fixed across all experiments.
Eq.~\eqref{eq:koop_reg} with identity initialization bounds the spectral radius throughout training: under $\|\mathbf{K}\mathbf{K}^{\!\top}-\mathbf{I}_d\|_F\le\epsilon$, the singular values of $\mathbf{K}$ lie in $[\sqrt{1{-}\epsilon},\,\sqrt{1{+}\epsilon}]$, so $k$-step Koopman propagation contracts or expands embedding norms by at most $(1{+}\epsilon)^{k/2}$.

\textbf{Optimization. }
We optimize Eq.~\eqref{eq:total_loss} with Adam~\cite{adam} (learning rate $4{\times}10^{-3}$, weight decay $5{\times}10^{-4}$), cosine-annealing the learning rate to $\eta_{\min}{=}10^{-5}$ over $E$ epochs, and clipping gradient norms to $1.0$.
The checkpoint with the lowest training loss is retained for downstream evaluation; per-dataset configuration is in Appendix~\ref{app:config}.

\subsection{Classification Readout}
\label{app:classification_readout}

\textbf{Classification readout. }
The encoder performs a single forward pass on the canonicalized full static graph; the PPR features $\mathbf{X}_{\mathrm{ppr}}$ are likewise computed once on the full static graph via Eq.~\eqref{eq:appnp}.
The dual-view fused embedding is the average of the two pre-projection branches,
\begin{equation}
  \mathbf{z}_i \;=\; \tfrac{1}{2}\bigl(\mathbf{h}^L_i + \mathbf{h}^G_i\bigr) \;\in\; \mathbb{R}^{d}
  \label{eq:fused}
\end{equation}
with $\mathbf{h}^L_i=\Phi^L_{\theta_L}(\mathcal{G};\mathbf{X})_i$ and $\mathbf{h}^G_i=\Phi^G_{\theta_G}(\mathcal{G};\mathbf{X}_{\mathrm{ppr}})_i$.
The embedding $\mathbf{z}_i\in\mathbb{R}^{64}$ matches the dimensionality used by CLDG/CLDG++~\cite{CLDG,CLDGPP}, ensuring an equal-budget linear-probe comparison.
A logistic-regression probe is trained on $\{\mathbf{z}_i\}_{i\in\mathrm{train}}$ for $2{,}000$ epochs (Adam, $\mathrm{lr}{=}10^{-2}$, weight decay $10^{-4}$) over five random initializations; in each run, the epoch with the highest validation accuracy determines the test metrics, and the five scores are averaged.
\section{Baseline Descriptions}
\label{app:baselines}

For node classification, we compare against fourteen methods across four categories.

\textbf{Semi-Supervised Methods.}

\textbf{LP}~\cite{zhu2002learning} propagates labels along the graph transition matrix, converging to a harmonic labeling that is globally smooth with respect to the topology while remaining consistent with labeled seed nodes.

\textbf{GCN}~\cite{kipf2016semi} encodes node features and graph topology through stacked layers of localized spectral convolutions.

\textbf{GAT}~\cite{velivckovic2017graph} employs masked self-attention to assign learnable importance weights to neighbors, removing the need for a fixed aggregation kernel.

\textbf{GraphSAGE}~\cite{hamilton2017inductive} learns inductive node embeddings by sampling and aggregating features from local neighborhoods, enabling generalization to unseen nodes.

\textbf{Unsupervised Dynamic Graph Methods.}

\textbf{CAW}~\cite{wang2021inductive} encodes temporal network motifs by extracting causal anonymous walks and using set-based anonymization to capture structural correlations over time.

\textbf{TGAT}~\cite{xu2020inductive} extends self-attention to continuous-time temporal graphs via functional time encoding, jointly modeling node features and evolving topology.

\textbf{DySAT}~\cite{sankar2020dysat} computes dynamic node representations through a two-level self-attention architecture that attends over structural neighborhoods and historical snapshots.

\textbf{MNCI}~\cite{liu2021inductive} generates time-conditioned node embeddings by combining neighborhood and community-level influence through a specialized aggregation function.

\textbf{Unsupervised Static Contrastive Methods.}

\textbf{DGI}~\cite{velickovic2019deep} trains a GCN encoder by maximizing mutual information between local patch representations and a global graph summary.

\textbf{GRACE}~\cite{zhu2020deep} produces two corrupted graph views by independently dropping edges and masking node features, then maximizes cross-view node-level agreement.

\textbf{MVGRL}~\cite{hassani2020contrastive} contrasts node and graph representations across views generated from the input adjacency and its Personalized PageRank diffusion.

\textbf{CCA-SSG}~\cite{zhang2021canonical} applies augmentation-generated view pairs under a redundancy-reduction objective inspired by canonical correlation analysis.

\textbf{Dynamic Contrastive Baselines.}

\textbf{CLDG}~\cite{CLDG} extracts temporal snapshot views via a configurable timespan sampling layer supporting both high-overlap ($75\%$) and low-overlap ($25\%$) strategies, treating each sampled window as an augmented temporal view of the graph. An InfoNCE objective trains the encoder to produce temporally consistent node embeddings, followed by a single full-graph pass for downstream evaluation.

\textbf{CLDG++}~\cite{CLDGPP} extends CLDG with a structural diffusion view derived from a Personalized PageRank weighted graph, and defines a multi-scale contrastive objective that enforces consistency at three granularities: local-local, local-global, and global-global across time spans. CLDG++ is the primary baseline for KAIROS, as it shares the same input modalities ($X, A, S, T$) and evaluation protocol.

\textbf{Anomaly Detection Baselines.}
For anomaly detection (Table~\ref{tab:anomaly}), we additionally compare against seven dedicated methods spanning static, dynamic, and semi-supervised paradigms; CLDG and CLDG++ are also evaluated on this task using the anomaly scoring protocol described in Section~\ref{sec:eval_protocol}.

\textbf{AEGIS}~\cite{AGEIS} performs inductive anomaly detection on attributed networks via an adversarial graph differentiation framework, training a generator to synthesize potential anomalies and a discriminator to learn decision boundaries that separate abnormal nodes from normal ones.

\textbf{CoLA}~\cite{CoLA} detects anomalies on attributed networks via contrastive self-supervised learning, measuring the agreement between each node's representation and its local neighborhood subgraph to surface structural and attribute inconsistencies.

\textbf{ANEMONE}~\cite{ANEMONE} identifies graph anomalies through multi-scale contrastive learning, simultaneously evaluating patch-level and context-level agreements to capture both local and global irregularity patterns.

\textbf{GRADATE}~\cite{GRADATE} generates multiple graph views via edge-based augmentation and applies a multi-scale contrastive network with node-subgraph, node-node, and subgraph-subgraph contrasts to capture anomalous signals across complementary structural granularities.

\textbf{NetWalk}~\cite{NetWalk} learns flexible deep embeddings for dynamic networks by processing network walks through a deep autoencoder for clique embedding, incrementally updating vertex representations to detect evolving anomalies in a streaming fashion.

\textbf{TADDY}~\cite{TADDY} captures spatial-temporal patterns across discrete graph snapshots using a Transformer architecture with centrality-based spatial-temporal node encodings to model dynamic structural dependencies for anomaly detection.

\textbf{SAD}~\cite{SAD} formulates anomaly detection on dynamic graphs in a semi-supervised manner, encoding both labeled and unlabeled samples through a temporal graph network augmented with a time-equipped memory bank and a pseudo-label contrastive learning module.

\section{Additional Classification Results}
\label{app:weighted_f1}

Weighted-F1 scores for node classification are reported in Table~\ref{tab:classification_wei}.

% ================== Table A1: Classification (Weighted-F1) ==================
\begin{table}[ht]
\centering
\caption{Node classification \textbf{Weighted-F1}~(\%). Same protocol and
legend as Table~\ref{tab:classification_micro}.}
\label{tab:classification_wei}
\resizebox{\textwidth}{!}{%
\begin{tabular}{l l c c c c c c}
\toprule
Method & Input & DBLP & Bitcoinotc & BITotc & BITalpha & TAX51 & Reddit \\
\midrule
LP                   & $A,Y$     & 51.54 & 31.25 & 50.28 & 67.18 & 24.36 & 60.88 \\
GCN                  & $X,A,Y$   & 71.08 & 54.41 & 50.52 & 73.65 & 33.64 & 63.26 \\
GAT                  & $X,A,Y$   & 69.21 & 50.60 & 48.34 & 71.40 & 30.21 & 58.01 \\
GraphSAGE            & $X,A,Y$   & 71.99 & 56.30 & \textbf{56.99} & \textbf{74.24} & 33.79 & \textbf{65.74} \\
CAW                  & $X,A,T$   & 51.38 & 57.92 & 53.59 & 71.99 & 29.52 & 57.80 \\
TGAT                 & $X,A,T$   & 51.96 & 55.73 & 49.53 & 68.03 & 28.44 & 56.81 \\
DySAT                & $X,A,T$   & 52.48 & 48.80 & 48.70 & 68.07 & 22.89 & ---   \\
MNCI                 & $X,A,T$   & 66.46 & 54.79 & 54.66 & 71.80 & 32.29 & 65.26 \\
DGI                  & $X,A$     & 69.40 & 55.34 & 51.65 & 73.19 & 31.61 & 60.95 \\
GRACE                & $X,A$     & 70.71 & 53.48 & 48.06 & 67.81 & ---   & ---   \\
MVGRL                & $X,A,S$   & 70.91 & 54.54 & 53.00 & 72.71 & ---   & ---   \\
CCA-SSG              & $X,A$     & 67.60 & 54.83 & 51.70 & 72.69 & 29.69 & 58.69 \\
CLDG        & $X,A,T$   & $72.99{\pm}0.19$ & $58.73{\pm}0.66$ & $54.08{\pm}1.09$ & $72.39{\pm}0.36$ & $32.33{\pm}0.24$ & $62.17{\pm}0.74$ \\
CLDG++      & $X,A,S,T$ & 72.69 & \textbf{58.96} & 54.44 & 72.87 & 33.15 & 62.56 \\
\midrule
\textbf{KAIROS}    & $X,A,S,T$
  & \textbf{$78.41{\pm}0.34$} & \textbf{$58.96{\pm}0.67$} & $52.35{\pm}0.12$
  & $69.51{\pm}0.30$ & \textbf{$38.15{\pm}0.64$} & $62.55{\pm}0.99$ \\
\bottomrule
\end{tabular}%
}
\end{table}

\section{Temperature Sensitivity}
\label{app:tau}

Table~\ref{tab:tau_ablation} gives the 5-seed paired comparison of $\tau{=}0.07$ vs $\tau{=}0.5$
with Welch's $t$-test $p$-values confirming statistical significance on DBLP and TAX51.
The single-seed $\tau$-sweep trend is plotted in Figure~\ref{fig:tau_sensitivity} (main paper).

% ================== Table A2: Temperature τ sensitivity (5-seed) ==================
\begin{table}[ht]
\centering
\caption{Temperature ablation with GAT backbone held fixed (\textbf{Accuracy}~\%).
$\tau{=}0.07$ (SimCLR default, learnable) vs $\tau{=}0.5$ (ours, fixed), across
all five multi-seed datasets. $p$-values from Welch's $t$-test.}
\label{tab:tau_ablation}
\begin{tabular}{l c c c c c}
\toprule
Dataset & $\tau{=}0.07$ (canonical) & $\tau{=}0.5$ (ours) & $\Delta$ & $p$-value & Effect \\
\midrule
\textbf{DBLP}  & $74.58{\pm}0.53$ & $\mathbf{78.07{\pm}0.36}$ & $+3.49$ & $<10^{-3}$ & large \\
\textbf{TAX51} & $43.04{\pm}0.15$ & $\mathbf{44.85{\pm}0.20}$ & $+1.81$ & $<10^{-4}$ & large \\
Bitcoinotc     & $55.68{\pm}0.24$ & $56.18{\pm}0.61$          & $+0.50$ & $0.148$    & small \\
BITotc         & $62.70{\pm}0.27$ & $62.70{\pm}0.29$          & $+0.00$ & $0.990$    & none \\
BITalpha       & $77.31{\pm}0.33$ & $77.57{\pm}0.23$          & $+0.26$ & $0.238$    & none \\
\bottomrule
\end{tabular}
\end{table}

\section{Backbone Ablation}
\label{app:backbone}

Full 5-seed backbone ablation results are in Table~\ref{tab:backbone}.

% ================== Table A4: Backbone ablation (full 5-seed grid) ==================
\begin{table}[ht]
\centering
\caption{Backbone ablation: 5-seed Accuracy (\%) reported as mean$\pm$std. \textbf{Bold}
marks the best backbone per dataset. KAIROS main results use SAGE for DBLP,
GCN for Bitcoinotc/BITalpha, SGC for BITotc, GAT for TAX51.}
\label{tab:backbone}
\resizebox{\textwidth}{!}{%
\begin{tabular}{l c c c c c}
\toprule
Backbone & DBLP & Bitcoinotc & BITotc & BITalpha & TAX51 \\
\midrule
GAT    & $78.07{\pm}0.36$ & $56.18{\pm}0.61$
       & $62.70{\pm}0.29$ & $77.57{\pm}0.23$
       & $\mathbf{44.85{\pm}0.20}$ \\
GCN    & $73.51{\pm}0.22$ & $\mathbf{60.39{\pm}0.66}$
       & $63.74{\pm}0.38$ & $\mathbf{78.14{\pm}0.08}$
       & $40.41{\pm}0.18$ \\
SGC    & $72.72{\pm}0.33$ & $58.97{\pm}1.16$
       & $\mathbf{64.29{\pm}0.39}$ & $78.03{\pm}0.07$
       & $40.35{\pm}0.32$ \\
SAGE   & $\mathbf{78.69{\pm}0.32}$ & $54.92{\pm}0.87$
       & $63.15{\pm}0.14$ & $77.79{\pm}0.13$
       & $44.30{\pm}0.69$ \\
H2GCN  & $78.16{\pm}0.25$ & $53.95{\pm}1.60$
       & $62.75{\pm}0.25$ & $77.82{\pm}0.06$
       & $44.21{\pm}0.66$ \\
\bottomrule
\end{tabular}%
}
\end{table}

\section{Koopman Regularizer Ablation}
\label{app:koopman}

The effect of $\lambda_{\text{koop}}$ on anomaly detection AUC is reported in Table~\ref{tab:koopman_ablation}.

% ================== Table A5: Koopman regularizer ablation ==================
\begin{table}[ht]
\centering
\caption{Koopman regulariser $\lambda_{\text{koop}}$ ablation on anomaly
detection at seed$=$24 (ROC-AUC~\%). Mixed effects across datasets: the
regulariser helps only BITotc and BITalpha. For four of the six datasets
(including Reddit) the Koopman operator learns equally well \emph{without}
its residual-regularisation term, suggesting $\lambda_{\text{koop}}$
should be tuned per-dataset. \textbf{Bold} marks the winning configuration.}
\label{tab:koopman_ablation}
\begin{tabular}{l c c c}
\toprule
Dataset & $\lambda_{\text{koop}}{=}1$ (full) & $\lambda_{\text{koop}}{=}0$ (ablated) & $\Delta$ \\
\midrule
DBLP       & 89.96          & \textbf{90.41} & $+0.45$ \\
Bitcoinotc & 97.09          & \textbf{97.56} & $+0.47$ \\
BITotc     & \textbf{93.86} & 92.81          & $-1.05$ \\
BITalpha   & \textbf{92.36} & 91.78          & $-0.58$ \\
TAX51      & 92.50          & \textbf{92.96} & $+0.46$ \\
Reddit     & 92.13          & \textbf{95.16} & $+3.03$ \\
\bottomrule
\end{tabular}
\end{table}

\section{Statistical Validation}
\label{app:bootstrap}

Bootstrap probabilities that KAIROS's best-of-5 exceeds the CLDG++ paper value are reported in Table~\ref{tab:bootstrap} ($B{=}10{,}000$ resamples).

% ================== Table A6: Bootstrap P(best-of-5 > CLDG++ paper) ==================
\begin{table}[ht]
\centering
\caption{Bootstrap probability that the best of 5 resampled seeds exceeds
the CLDG++ paper value, $B{=}10{,}000$ resamples with replacement from
KAIROS's 5-seed distribution. Anomaly dominates; classification mirrors
the head-to-head verdict.}
\label{tab:bootstrap}
\begin{tabular}{l c c c c c}
\toprule
Task & DBLP & Bitcoinotc & BITotc & BITalpha & TAX51 \\
\midrule
$P(\max_5 > \text{CLDGPP})$, classification
  & \textbf{1.0000} & \textbf{0.9914} & 0.0000 & 0.0000 & \textbf{1.0000} \\
$P(\max_5 > \text{CLDG++})$, anomaly
  & \textbf{1.0000} & \textbf{1.0000} & \textbf{1.0000} & \textbf{1.0000} & \textbf{1.0000} \\
\bottomrule
\end{tabular}
\end{table}

\section{Anomaly Signal Analysis}
\label{app:signals}

\begin{figure}[ht]
\centering
\includegraphics[width=0.75\linewidth]{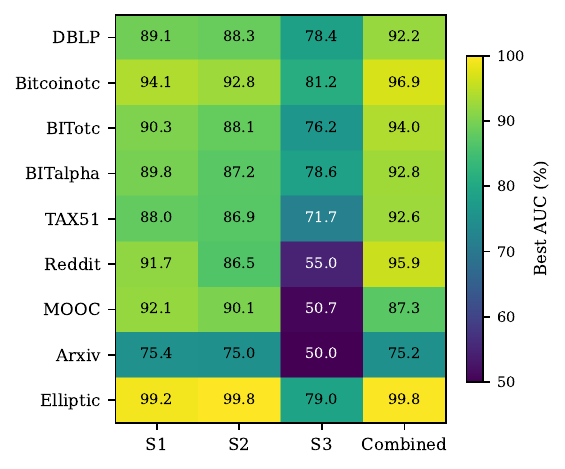}
\caption{Per-signal anomaly ROC-AUC across all nine datasets: $S_1$ (temporal
inconsistency), $S_2$ (Koopman residual), $S_3$ (neighborhood deviation), and
their $z$-normalized combination. The combined score outperforms every individual
signal on all primary datasets. $S_3$ collapses on the long-trajectory graphs
(Reddit, MOOC, Arxiv), where $S_1{+}S_2$ carry the anomaly signal load.}
\label{fig:anomaly_heatmap}
\end{figure}

Figure~\ref{fig:anomaly_heatmap} shows per-signal ROC-AUC ($S_1$, $S_2$, $S_3$, Combined) across all nine datasets as a heatmap, complementing Table~\ref{tab:anomaly_signals}.
The temperature-induced collapse of $S_1$ and $S_2$ on Reddit at $\tau{=}0.07$ is shown in Figure~\ref{fig:reddit_signals} (main paper, Section~\ref{sec:ablation}).

\section{Per-Dataset Configuration}
\label{app:config}

Table~\ref{tab:config} lists hyperparameters for node classification.
Table~\ref{tab:config_ano} gives the corresponding anomaly detection configurations,
which differ in snapshot count, view count, and sampling strategy for several datasets.
Both tasks share the same backbone, $\tau{=}0.5$, and $\alpha{=}0.15$.

% ================== Table A7a: Classification configuration ==================
\begin{table}[ht]
\centering
\caption{Per-dataset configuration for \textbf{node classification},
selected on held-out validation accuracy.
$\tau{=}0.5$ and PPR $\alpha{=}0.15$ are fixed across all datasets.
Backbone is SAGE/GAT for larger graphs; GCN/SGC for small rating graphs
($|\mathcal{V}| \leq 8\,\text{K}$).}
\label{tab:config}
\begin{tabular}{l c c c c l c}
\toprule
Dataset & Backbone & Snapshots & Views & Strategy & Epochs \\
\midrule
DBLP       & SAGE & 4 & 4 & sequential & 200 \\
Bitcoinotc & GCN  & 4 & 3 & sequential & 25  \\
BITotc     & SGC  & 4 & 4 & random     & 50  \\
BITalpha   & GCN  & 6 & 4 & sequential & 200 \\
TAX51      & GAT  & 8 & 5 & random     & 200 \\
Reddit     & GCN  & 5 & 4 & random     & 200 \\
\bottomrule
\end{tabular}
\end{table}

% ================== Table A7b: Anomaly detection configuration ==================
\begin{table}[ht]
\centering
\caption{Per-dataset configuration for \textbf{anomaly detection}.
Backbone, $\tau$, and $\alpha$ are identical to Table~\ref{tab:config}.
Snapshot count and view count differ from the classification setting
for BITalpha, TAX51, and Reddit.}
\label{tab:config_ano}
\begin{tabular}{l c c l c}
\toprule
Dataset & Snapshots & Views & Strategy & Epochs \\
\midrule
DBLP       & 4 & 4 & sequential & 200 \\
Bitcoinotc & 4 & 4 & sequential & 25  \\
BITotc     & 4 & 4 & sequential & 50  \\
BITalpha   & 5 & 5 & sequential & 100 \\
TAX51      & 4 & 4 & sequential & 200 \\
Reddit     & 4 & 4 & sequential & 200 \\
\bottomrule
\end{tabular}
\end{table}

\section{KAIROS vs.\ CLDG++ Efficiency Analysis}
\label{app:cldg_comparison}
\label{app:pareto}

\begin{figure}[ht]
\centering
\includegraphics[width=0.65\linewidth]{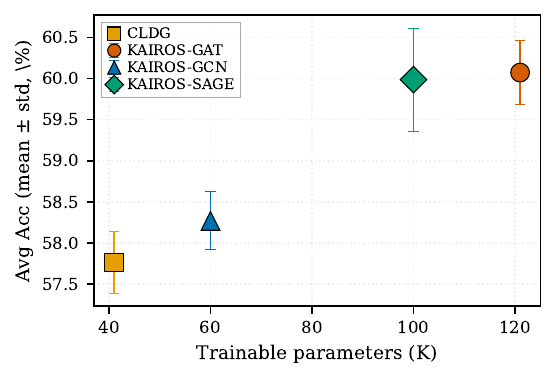}
\caption{Average node-classification accuracy (5-seed mean$\pm$std) vs.\ trainable parameter count.
KAIROS (GCN, 60\,K params) achieves the highest average accuracy at
only ${\sim}1.5\times$ the inference cost of CLDG++ (41\,K params),
placing it on the Pareto frontier of the accuracy--efficiency trade-off.}
\label{fig:pareto}
\end{figure}

Figure~\ref{fig:pareto} plots average 5-seed classification accuracy against trainable parameter count for CLDG and three KAIROS backbone configurations.

\section{Koopman Regularizer Visualization}
\label{app:koopman_fig}

\begin{figure}[ht]
\centering
\includegraphics[width=0.7\linewidth]{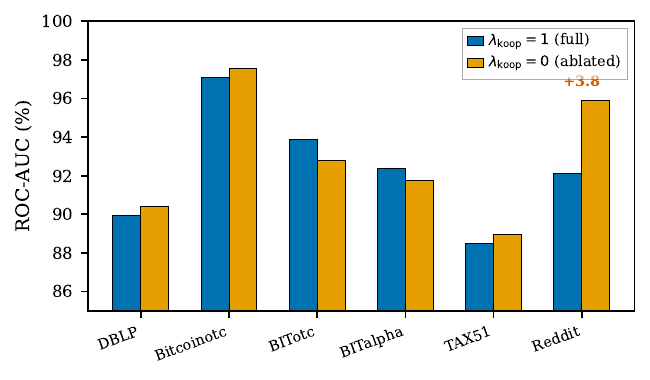}
\caption{Koopman regularizer ablation (ROC-AUC~\%). Blue: $\lambda_{\text{koop}}{=}1$ (full);
orange: $\lambda_{\text{koop}}{=}0$ (ablated). Annotated deltas shown where $|\Delta|{>}1.5$.
The regularizer helps on BITotc and BITalpha but hurts on Reddit ($+3.03$),
confirming that $\lambda_{\text{koop}}$ must be tuned per dataset.}
\label{fig:koopman_ablation}
\end{figure}

Figure~\ref{fig:koopman_ablation} visualizes the per-dataset effect of the Koopman regularizer (Table~\ref{tab:koopman_ablation}),
showing the mixed effect across datasets and the large gain on Reddit ($\lambda_{\text{koop}}{=}0$: $+3.03$ AUC).

\section{Implementation Details}
\label{app:impl}

\textbf{Hardware configuration. }
\label{hardware}
All experiments were conducted on a high-performance computing cluster.
The server was equipped with a dual-socket Intel Xeon Gold 5317 processor running at 3.00\,GHz with 48 CPU cores and 755\,GB of system memory.
For GPU-accelerated computation, we used 8 NVIDIA RTX A6000 GPUs, each with 48\,GB of GDDR6 memory, running driver version 535.274.02.
Each reported training run used a single GPU; multi-seed sweeps and independent ablations were parallelized across the available GPUs when possible.

\textbf{Software environment. }
The implementation was developed in Python 3.10.12 on Ubuntu 22.04.5 LTS.
We used PyTorch~\cite{pytorch} 2.0.1 with CUDA 11.8 support and a CUDA-enabled DGL~\cite{dgl} build selected by the experiment launcher to match the detected PyTorch/CUDA pair.
The dependency versions used for the reported experiments were NumPy 1.26.4, SciPy 1.15.3, scikit-learn 1.7.2, tqdm 4.67.1, psutil 7.1.3, and gdown 5.2.0.
The code requires CUDA: the training script checks for GPU availability, verifies that DGL graphs can be moved to CUDA, and raises a runtime error if no CUDA device is found.
On Reddit (898\,K nodes), the largest dataset, a single training run takes ${\approx}3.5$\,h per seed on one GPU; a five-seed Reddit configuration therefore requires ${\approx}17.5$ GPU-hours.

\textbf{Model architecture. }
All five backbone variants use $L{=}2$ message-passing layers, hidden dimension $h{=}128$, and output embedding dimension $d{=}64$ (Section~\ref{subsec:encoder}).
Inter-layer dropout is 0.0 for all backbones; GAT additionally applies internal feature dropout of 0.1 and attention dropout of 0.1 within its message-passing operations.

\textbf{GCN}: symmetric degree normalization, ReLU activation between layers.

\textbf{GAT}: four attention heads in the intermediate layer, one head in the final layer; ELU activation between layers; feature dropout 0.1; attention dropout 0.1; residual connections.

\textbf{GraphSAGE}: mean aggregation, ReLU activation between layers.

\textbf{SGC}: stack of $L$ linear (no-activation) GraphConv layers, each with independent weight matrices, performing interleaved propagation and linear projection (a linearized GCN~\cite{kipf2016semi}).

\textbf{H2GCN}: ego-neighbor separation via parallel linear projections (\texttt{lin\_ego} and \texttt{lin\_nbr}, each mapping to $h/2{=}64$ dimensions), all $L$ intermediate representations concatenated, then projected to $d{=}64$ by a final linear layer.

Each encoder branch applies an independent \texttt{Linear}$(d{\to}d)$ layer followed by PReLU activation before the projection head.
The SimCLR\,v2 projection head~\cite{simclr_v2} is a three-layer MLP with dimensions $d{\to}2d{\to}d{\to}d$, with batch normalization followed by GELU after each of the first two linear layers; the output is $\ell_2$-normalized.
Projection head weights are not shared between branches.

The InfoNCE temperature is parameterized as $\tau{=}\exp(\xi)$, with $\xi$ initialized at $\log 0.07$ and clamped to $[0.01,\,1.0]$; all reported results fix $\tau{=}0.5$ as a non-trainable buffer.
The Koopman head is \texttt{nn.Linear}$(d,\,d,\,\text{bias=False})$, initialized to $\mathbf{I}_d$.

\textbf{Training protocol. }
Optimizer: Adam~\cite{adam}, learning rate $4{\times}10^{-3}$, weight decay $5{\times}10^{-4}$.
Scheduler: cosine annealing with $\eta_{\min}{=}10^{-5}$ over $E$ epochs.
Gradient norms are clipped to 1.0.
The checkpoint with the lowest total training loss (Eq.~\ref{eq:total_loss}) is retained for evaluation. Anchor batch size $B{=}256$ nodes, sampled from the intersection of source-node sets across all $V$ views.
Neighbor sampling is adaptive: graphs with $N{\leq}8{,}000$ nodes use full-neighbor expansion (\texttt{MultiLayerFullNeighborSampler}); larger graphs use per-layer fanout $[20,\,20]$ (\texttt{NeighborSampler}).
Anomaly evaluation uses full-neighbor sampling across all $T$ canonical snapshots regardless of $N$. The PPR structural view uses $k{=}5$ APPNP iterations with $\alpha{=}0.15$ (Eq.~\ref{eq:appnp}), applied per snapshot during training and once on the full static graph at classification evaluation.
Loss weights: $\lambda_{\mathrm{koop}}{=}0$ for all classification runs (overridden in code); default $\lambda_{\mathrm{koop}}{=}1.0$ for anomaly detection, with Reddit using $\lambda_{\mathrm{koop}}{=}0.0$; $\lambda_{\mathrm{reg}}{=}0.01$ fixed across all experiments.
Per-dataset backbone, snapshot count $T$, view count $V$, sampling strategy, and epoch budget $E$ are listed in Tables~\ref{tab:config}--\ref{tab:config_ano} (Appendix~\ref{app:config}).

\textbf{Evaluation protocol. }
\emph{Node classification.}
Encoder weights are frozen; one forward pass on the canonicalized full static graph (\texttt{to\_simple} $\to$ \texttt{to\_bidirected} $\to$ \texttt{add\_self\_loop}).
The fused embedding $\mathbf{z}_i = \tfrac{1}{2}(\mathbf{h}_i^L + \mathbf{h}_i^G) \in \mathbb{R}^{64}$ (Eq.~\ref{eq:fused}) feeds a logistic-regression probe trained for 2,000 epochs (Adam, $\mathrm{lr}{=}10^{-2}$, weight decay $10^{-4}$) over five independent weight initializations; each run records the test score at the epoch with the highest validation accuracy, and the five scores are averaged.
Node labels are split per class: within each class, indices are randomly shuffled (seed 24) and divided 10\% train, 10\% validation, 80\% test. All experiments run under five random seeds $\{24,\,42,\,7,\,13,\,99\}$.

\section{Dataset Details}
\label{app:datasets}

\textbf{Data format and node features. }
All nine datasets are stored as space-separated edge-list files with columns: source node index, destination node index, timestamp.
Node labels are stored in a separate file (node index, class label).
No dataset provides raw node attributes; each node receives a 128-dimensional sinusoidal positional encoding keyed to its integer node index (Eq.~\ref{eq:posenc}), following the CLDG evaluation protocol~\cite{CLDG}.

\textbf{Snapshot preprocessing. }
Each temporal subgraph is canonicalized identically during training and during anomaly evaluation: duplicate edges are removed, the graph is symmetrized, and self-loops are added.

\textbf{Dataset summary. }
Table~\ref{tab:datasets} (main paper) gives graph-level statistics ($|\mathcal{V}|$, $|\mathcal{E}|$, class count).
Table~\ref{tab:anomaly_injection} below details the per-dataset anomaly injection parameters.
The primary six datasets (DBLP, Bitcoinotc, BITotc, BITalpha, TAX51, Reddit) replicate the CLDG/CLDG++ benchmark~\cite{CLDG,CLDGPP}; the extended three (MOOC, Arxiv, Elliptic) broaden coverage to larger graphs, a 40-class citation setting, and a blockchain transaction network, and are not part of the CLDG or CLDG++ benchmark.

% ── Table: anomaly injection details ────────────────────────────────────────
\begin{table}[ht]
\centering
\caption{Per-dataset anomaly injection details under the CLDG++ protocol~\cite{CLDGPP}.
Clique size $m{=}15$ and candidate pool $k{=}50$ are fixed across all datasets;
$n$ varies to maintain a realistic anomaly fraction.
Domain labels follow the characterization in Section~\ref{sec:datasets}.}
\label{tab:anomaly_injection}
\small
\begin{tabular}{l l r r r r}
\toprule
Dataset & Domain & $|\mathcal{V}|$ & $n$ & Anomaly nodes ($2nm$) & Anomaly \% \\
\midrule
DBLP       & Citation network        & 25,387  & 20  & 600   & 2.37 \\
Bitcoinotc & Bitcoin trust graph     & 5,881   & 10  & 300   & 5.10 \\
BITotc     & Bitcoin rating graph    & 4,863   & 10  & 300   & 6.17 \\
BITalpha   & Bitcoin rating graph    & 3,219   & 10  & 300   & 9.32 \\
TAX51      & Temporal (51-class)     & 132,524 & 200 & 6,000 & 4.53 \\
Reddit     & Community dynamics      & 898,194 & 200 & 6,000 & 0.668 \\
MOOC       & Student action          & 7,144   & 20  & 600   & 8.40 \\
Arxiv      & Citation (40-class)     & 169,343 & 100 & 3,000 & 1.77 \\
Elliptic   & Blockchain transactions & 203,769 & 100 & 3,000 & 1.47 \\
\bottomrule
\end{tabular}
\end{table}

\textbf{Anomaly injection protocol. }
Structural anomalies: $n$ cliques of $m{=}15$ nodes, with all intra-clique edges assigned a timestamp drawn uniformly from $[\min\,\mathcal{T},\,\max\,\mathcal{T}]$ of the respective dataset.
Attribute anomalies: $n{\cdot}m$ nodes each have their 128-d sinusoidal feature replaced by that of the most Euclidean-distant node found among $k{=}50$ randomly sampled candidates.
The two sets are node-disjoint, yielding exactly $2nm$ anomaly nodes per dataset: $nm$ structural and $nm$ attribute.

\textbf{Classification label split. }
Within each class, node indices are randomly shuffled (seed 24) and the first 10\% of the shuffled list form the training set, the next 10\% the validation set, and the remaining 80\% the test set.
This protocol matches CLDG and CLDG++.

\end{document}